\documentclass{article}

\usepackage[main, final]{neurips_2025}
\usepackage[utf8]{inputenc} % allow utf-8 input
\usepackage[T1]{fontenc}    % use 8-bit T1 fonts
\usepackage{hyperref}       % hyperlinks
\usepackage{url}            % simple URL typesetting
\usepackage{booktabs}       % professional-quality tables
\usepackage{amsfonts}       % blackboard math symbols
\usepackage{nicefrac}       % compact symbols for 1/2, etc.
\usepackage{microtype}      % microtypography
\usepackage{xcolor}         % colors
\usepackage{makecell}

\usepackage{graphicx}

\title{Cross-modal Associations in Vision and Language Models: Revisiting the Bouba-Kiki Effect}

\author{Tom Kouwenhoven$^{\ast}$, Kiana Shahrasbi, Tessa Verhoef\thanks{Equal contribution} \\
        Leiden Institute of Advanced Computer Science \\ Leiden University,
The Netherlands \\ \texttt{\{t.kouwenhoven, k.shahrasbi, t.verhoef\}}@liacs.leidenuniv.nl}

\begin{document}

\maketitle

\begin{abstract}
Recent advances in multimodal models have raised questions about whether vision-and-language models (VLMs) integrate cross-modal information in ways that reflect human cognition.
One well-studied test case in this domain is the bouba-kiki effect, where humans reliably associate pseudowords like `bouba' with round shapes and `kiki' with jagged ones.
Given the mixed evidence found in prior studies for this effect in VLMs, we present a comprehensive re-evaluation focused on two variants of CLIP, ResNet and Vision Transformer (ViT), given their centrality in many state-of-the-art VLMs.
We apply two complementary methods closely modelled after human experiments: a prompt-based evaluation that uses probabilities as a measure of model preference, and we use Grad-CAM as a novel approach to interpret visual attention in shape-word matching tasks.
Our findings show that these model variants do not consistently exhibit the bouba-kiki effect.
While ResNet shows a preference for round shapes, overall performance across both model variants lacks the expected associations.
Moreover, direct comparison with prior human data on the same task shows that the models' responses fall markedly short of the robust, modality-integrated behaviour characteristic of human cognition.
These results contribute to the ongoing debate about the extent to which VLMs truly understand cross-modal concepts, highlighting limitations in their internal representations and alignment with human intuitions.
\end{abstract}

%TL:DR - We re-evaluate whether vision-and-language models (variants of CLIP) exhibit the human-like bouba-kiki effects, using two methods modelled after human experiments. Compared to humans, VLMs fall short in aligning cross-modal associations with human intuitions.

\section{Introduction}
Recent advances in multimodal models that integrate vision and language have brought artificial intelligence a step closer to understanding the world in ways that resemble human experience and cognition. 
These models, which learn from vast amounts of paired visual and textual data, have demonstrated impressive capabilities in tasks such as image captioning, visual question answering, and cross-modal retrieval \citep{radford2021learning,li2022blip,li2023blip2}. 
Yet it remains unclear whether these models integrate visual and linguistic information in ways that parallel human cognitive processes. 
In this paper, we investigate whether VLMs exhibit human-like patterns of association between abstract visual shapes and unfamiliar words. 
Particularly, we focus on one of the most widely studied cross-modal test cases in human cognition, the bouba-kiki effect, in which people consistently associate pseudowords like `bouba' with round shapes and those like `kiki' with jagged shapes \citep{ramachandran2001synaesthesia, maurer2006boubas, cwiek2022acrossculture}. 
These associations have recently become a test case for evaluating whether language models trained on large-scale data show similar patterns \citep{alper2024kiki, verhoef-etal-2024-kiki, loakman-etal-2024-ears, Iida2024iconicityVLMs}. 
Results, however, have been mixed. 
While \citet{alper2024kiki} found overwhelming evidence for a bouba-kiki effect in CLIP and Stable Diffusion, other studies have found less convincing patterns across different VLMs, testing methods and datasets \citep{verhoef-etal-2024-kiki, loakman-etal-2024-ears, Iida2024iconicityVLMs}.

This work therefore revisits this effect by thoroughly testing two versions of CLIP \citep{radford2021learning}: ResNet \citep{he2016deep} and ViT \citep{dosovitskiy2021an}.
We focus on CLIP because, among four models (CLIP, BLIP2, ViLT, and GPT-4o) tested, \cite{verhoef-etal-2024-kiki} found that CLIP demonstrated the most promising alignment with a human-like bouba-kiki effect. 
This aligns with previous work demonstrating that CLIP outperforms other models in capturing human-like decision patterns \citep{demircan2024evaluating}.
Perhaps most importantly, CLIP often acts as a foundational model in state-of-the-art VLMs (see \autoref{sec:methods}). 
If a `base' model does not exhibit human-like associations, it is difficult to imagine that a model using that base model as a backbone will show human-like preferences without explicit fine-tuning on relevant cross-modal associations. 
Unravelling how these `base' models represent cross-modal information additionally benefits our understanding of limitations in, for example, spatial reasoning \citep{Thrush2022Winoground}, (in-context) shape-colour biases in VLMs \citep{allen2025the}, and emergent communication setups \citep{kouwenhoven-etal-2024-curious}.
% Perhaps most importantly, given CLIP's contribution to many state-of-the-art VLMs, unravelling what and how these `base' models represent cross-modal information can additionally benefit our understanding of limitations in, for example, spatial reasoning \citep{Thrush2022Winoground} and (in-context) shape-colour biases in VLMs \citep{allen2025the}.

By probing these models' preferences for shape-word association in novel contexts, we aim to shed light on the nature of their internal representations and their alignment with human intuitions. 
% When differences of this kind exist, training models to develop human-like prior expectations or instilling them into models may help create a flexible, grounded understanding of language, resulting in conceptually fluent and natural human-machine interactions where machines intuitively understand what we mean, even in unfamiliar settings \citep{lake2017buildingmachines}. 
When these differ, training models to develop human-like prior expectations or instilling them into models may help and could even improve learning efficiency \citep{lake2017buildingmachines}.
Doing so is essential, since a flexible, grounded understanding of this kind could enable conceptually fluent, natural human-machine interactions, in which machines intuitively understand what we mean, even in unfamiliar settings.
% A lack thereof may hinder natural interactions between humans and machines \citep{verhoef-etal-2024-kiki}, especially when considering the dynamic and adaptive nature of human language. 
To this end, we employ two complementary methodological approaches that address the same cross-modal associations from different perspectives, thereby minimising the limitations of relying on a single evaluative framework.  
Despite this comprehensive strategy, we do not find compelling evidence for the bouba-kiki effect.
This work contributes in the following ways:

(1) We test for the bouba-kiki effect in VLMs in ways that are as close as possible to how it has been tested with humans in the past. This enables direct comparisons between our results and human findings, building on and expanding the work by \citet{verhoef-etal-2024-kiki}. 
Tested comprehensively, we find that models, compared to humans, do not make consistent cross-modal associations.

(2) We introduce a novel way to test models using Grad-CAM \citep{Selvaraju2019Grad-CAM}, a method from the model interpretability literature, to look more closely at visual processing in the bouba-kiki context. Strengthening the robustness of our results, it reveals that the models do not explicitly focus on bouba-kiki related shape-specific features.
%(3) Including both ViT and ResNet-based CLIP models in our analysis allows us to examine how architectural differences within the same training paradigm influence model behaviour. While none of these versions display human-like associations, ResNet seems to have a preference for matching round-sounding labels with round shapes.

%Perhaps:The findings are placed in context of various other works investigating multi-modal models 

\begin{figure}
    \centering
    \includegraphics[width=1\linewidth]{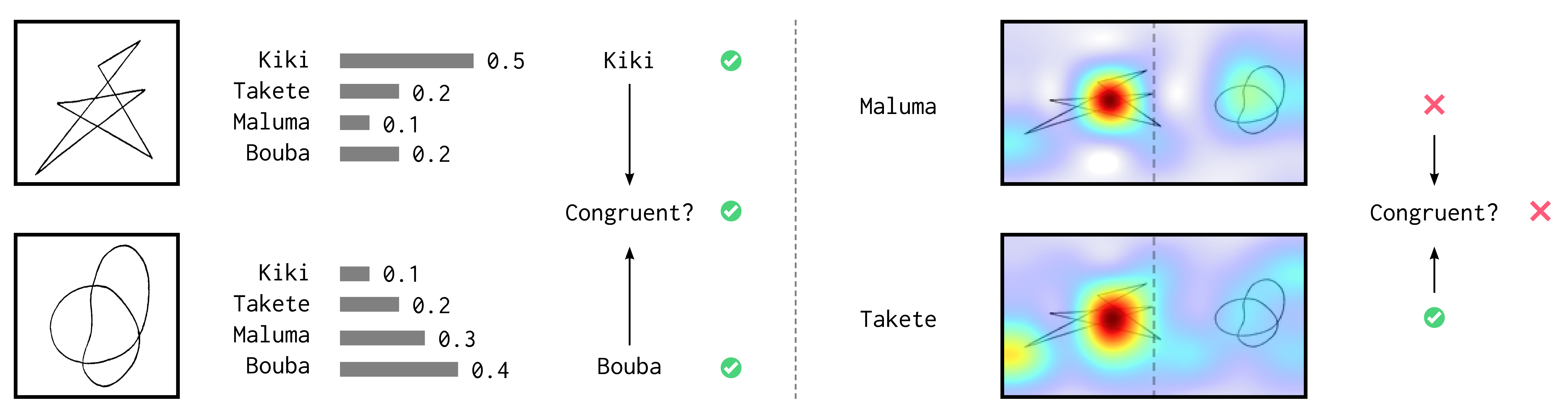}
    \caption{An overview of the two complementary methods used. On the left, we calculate the probabilities for each label across the four original pseudowords (note that the number of labels varies per label source) for each image shape and select the label with the highest probability (values are exemplary). On the right, we use concatenated image pairs and their labels as targets to calculate attention patterns with Grad-CAM and select the shape with the highest sum of attention.}
    \label{fig:method-overview}
\end{figure}

\section{Related work}
First, we discuss how prior work on cross-modal associations in human language and cognitive processing has shown that non-arbitrariness is both pervasive and affects how we learn, process and develop language. 
We then present previous studies that have tested for the bouba-kiki effect in VLMs.

\subsection{Cross-modal associations in human language}
Traditionally, it has been argued that mappings between words and their meanings are largely arbitrary.
\citet{hockett1960origin} uses the words `whale' and `microorganism' as an example: `whale' is a short word for a large animal, while `microorganism' is the reverse.
However, growing evidence from fields like cognitive science, language evolution and (sign language) linguistics suggests non-arbitrary form-meaning mappings are more widespread in human language than initially thought.
This suggests that it should be considered a general property of human language \citep{perniss2010iconicity}, shaped by cognitive mechanisms similar to those involved in the bouba-kiki effect.
Especially when looking beyond Indo-European languages, `iconic' mappings, where word forms seem to resemble their meanings, appear to play a significant role in many languages \citep{perniss2010iconicity,dingemanse2012advances,imai2014sound}. 
Some languages have specific word classes where characteristics of the meaning are mimicked or iconically represented in the word.
For example, Japanese ideophones allow speakers to depict sensory information through word forms, such as saying \textit{nuru nuru} when describing something as `slimy' and \textit{fuwa fuwa} when it is `fluffy' \citep{dingemanse2016sound}. 
Similarly, in Siwu, \textit{pimbilii} means `small belly', while \textit{pumbuluu} refers to `fat belly', and non-speakers of those languages can typically guess the meanings of such words \citep{dingemanse2016sound}. 
Even in languages not considered rich in sound symbolism, such as English and Spanish, vocabulary items from specific lexical categories, such as adjectives, are rated relatively high in iconicity \citep{perry2015iconicity}. 

Evidence for strong associations between speech sounds and particular meanings has been found in a broad sample of vocabularies from two-thirds of the world's languages \cite{blasi2016sound}.
Iconic mappings are not only widespread in the world's languages, they also help both children \cite{imai2008sound,perry2015iconicity,perry2018iconicity} and adults \cite{nielsen2012source} learn new words more easily.
They, moreover, allow us to communicate successfully even when a shared language is absent or existing vocabulary is insufficient, because cross-modal associations lay the groundwork for the negotiation of novel words and their meanings \citep{ramachandran2001synaesthesia,cuskley2013syn,imai2014sound}. 
Shared cross-modal representations enable people to instantly interpret unfamiliar words even on first exposure by directly linking sensory experiences with meaning. 
Indeed, experiments with humans that studied the influence of cross-modal preferences on the emergence of novel vocabularies show that iconic strategies are frequently adopted when word forms and meanings can be intuitively mapped, and they help to communicate successfully \citep{verhoef2015emergence,verhoef2016cognitive,verhoef2016iconicity,tamariz2018interactive}. 
An understanding of such mappings, however, requires human-like integration of multi-sensory information or an awareness of common cross-modal associations. 
While multimodal computational models have demonstrated remarkable capabilities across a range of tasks, the internal mechanisms through which these systems form and link representations remain opaque. 
In particular, it is still an open question whether these models integrate visual and linguistic information in ways that mirror human cognitive processes.

\subsection{Bouba-kiki effect in VLMs}
The bouba-kiki effect, as a specific example of visuo-linguistic processing, has been studied in VLMs before; however, the results so far seem to be conflicting.
One key study in this space is the innovative work by \citet{alper2024kiki}, who convincingly demonstrated that patterns aligning with the bouba-kiki effect are reflected in the embedding spaces of CLIP and Stable Diffusion. 
They used Stable Diffusion to generate images based on pseudowords that were carefully designed to reflect phonetic properties associated with sharp or round shapes.  
Specifically, they used CLIP’s text encoder to embed prompts containing either pseudowords or descriptive adjectives, while the images generated from pseudoword-based prompts were passed through CLIP’s vision encoder. 
This setup allowed both text and image representations to be analysed within the same multimodal embedding space. The embedding similarity between the pseudowords and the adjective or image representations was then used to compute geometric and phonetic scores, indicating alignment with sharp or round associations.
They concluded that their findings indicate strong evidence for the existence of cross-modal associations in VLMs. 
%[We could consider to add here some discussion about why this result is surprising given the explanations often mentioned for humans having this bias, which all involve an aspect of  situated real-world experience with a body and the environment, which these models don't have + other known limitations of current known VLMs in visual grounding]
Typical explanations given for the presence of a bouba-kiki effect in humans include experience with acoustics and articulation \citep{ramachandran2001synaesthesia, maurer2006shape, westbury2005implicit}, affective–semantic properties of human and non-human vocal communication \citep{nielsen2011sound}, or physical properties relating to audiovisual regularities in the environment \citep{fort2022resolving}.
This renders the conclusion by \citet{alper2024kiki} somewhat surprising, as these explanations all involve situated, real-world experience with a body and the environment, something these models entirely lack. 
Moreover, limitations of VLMs in visual grounding have been observed in many other domains \citep{Thrush2022Winoground, diwan-etal-2022-winoground, kamath-etal-2023-whats, jabri2016revisiting-vqa,goyal2019vqa, agrawal2018dont-just-assume, jones-etal-2024-multimodal}, suggesting that these models do not integrate textual and visual data in a manner that is human-like.
%For example, these models tend to answer "yes" to binary questions without basing their answers on the image. The work by \cite{goyal2019vqa} demonstrates that even when these models are trained on a balanced dataset, they continue to struggle to connect textual input with subtle visual details.
% Collectively, these examples suggest that even if VLMs can extract sound-symbolic information from the texts they have been trained on, they are likely to have difficulty associating this information with visual features.
Nonetheless, this work presented an innovative method for testing the bouba-kiki effect in VLMs. 
It convincingly demonstrated that VLMs encode relationships between word forms and semantic concepts related to roundness and jaggedness.
As described previously, these associations are indeed abundantly present in human languages, and prior work with text-only language models has also shown that these models can detect such regularities \citep{abramova-fernandez-2016-questioning, pimentel-etal-2019-meaning, devarda2022Cross-Modal, marklová2025iconicitylargelanguagemodels}.

%A similar point is made by \cite{devarda2022Cross-Modal}, who argued that some sound–meaning associations are unimodal in nature, and may come from the structure of language itself, rather than from cross-modal associations. In their study, they showed that strong form–meaning correspondences exist across different languages, even when visual input is not involved.

Interestingly though, \cite{Iida2024iconicityVLMs} replicated \citeauthor{alper2024kiki}'s experiments for Japanese and found that Japanese VLMs did not exhibit the expected bouba-kiki effect, even though Japanese is a language rich in sound-symbolism \citep{dingemanse2012advances}, and speakers of Japanese display the strongest bouba-kiki effect compared to speakers of 25 other languages in a study by \cite{cwiek2022acrossculture}. 
These findings suggest that \citeauthor{alper2024kiki}'s method may not capture true sensory mappings, but instead detects regularities between word forms and meanings that are language-specific rather than universal.
For example, sounds like \emph{/p/}, which are typically linked to sharpness according to the bouba-kiki effect, are strongly associated with roundness in ideophonic Japanese words, like \emph{pocha-pocha} `chubby’, \emph{puyo-puyo} `fat’, and \emph{puku-puku} `puffing up’ \citep{Iida2024iconicityVLMs}.
This suggests that the method used to disambiguate sharp and round pseudowords and images may pick up on relationships between semantic concepts and word forms—being heavily entangled with the choice of ground-truth adjectives—rather than capturing true sensory mappings in languages. 
Moreover, the vectors used to assign a geometric or phonetic score to a pseudoword or image must be sufficiently dissimilar, for which \citet{Iida2024iconicityVLMs} reported that this was not the case.
The contradictory findings between Japanese and English VLMs highlight the need for a different approach that aligns with human experiments on cross-modal associations.

A key ingredient of human bouba-kiki experiments is that tests are centred around specifically designed pairs of visual images that minimally differ, but with one more rounded and one more jagged version, as shown in \autoref{fig:method-overview}.
For example, \citet{maurer2006boubas} presented a pair of jagged/round images along with a pair of bouba/kiki-like words and participants were asked to pair them up in the most fitting way. 
Other studies employed more stringent tests, as in \citet{cwiek2022acrossculture}, where only the images were presented side by side, and participants had to choose the best-fitting image after listening to only one of the spoken words at a time.
Finally, \citet{nielsen2013parsing} presented single images and asked participants to generate novel pseudowords to match the shapes. 
In all of these cases, humans exhibit a strong bouba-kiki effect. 
Inspired by these studies, \citet{verhoef-etal-2024-kiki} used images from existing human experiments and explored four prominent VLMs on carefully designed image-to-word matching tasks.
They directly used model probabilities generated by VLMs to match specific pseudowords with images as a measure of preference and found limited evidence for the bouba-kiki effect. 
Two out of four models (CLIP and GPT-4o) exhibit moderate alignment with human-like associations, but only in some of the tests they conducted, and not consistently.
The study concludes that cross-modal associations in VLMs are highly dependent on factors such as model architecture, training data, and the specific test used. 

Others have also investigated the bouba-kiki effect and other cross-modal associations, such as a relation between perceived size and vowels \citep{loakman-etal-2024-ears} and understanding shitsukan terms (a Japanese concept that captures the sensory essence of an object) \citep{shiono-etal-2025-evaluating}.
However, none of these consistently find a resemblance between the human and model associations. 
\citet{tseng-etal-2024-measuring} used the embedding method from \citeauthor{alper2024kiki} to test sensitivity to sound-symbolic associations in audio-visual models and report that these models capture sound-meaning connections akin to human language processing. 
Yet, this method again seems to rely on the relationships between semantic concepts and word forms, as was mentioned before. 
Given the conflicting evidence in this domain, we revisit the bouba-kiki effect through a thorough investigation using two versions of CLIP and a wide variety of prompts. 
This approach introduces a novel method for testing it using visual interpretability methods and directly compares the results with similar findings from human studies.

\section{Methodology}\label{sec:methods}
Cross-modal associations are assessed by prompting two different versions of CLIP \citep{radford2021learning}, specifically, a ResNet-50 and a ViT version. 
We deem this as reasonable since state-of-the-art VLMs such as Molmo \citep{deitke2024molmopixmoopenweights}, LLaVA \citep{liu2023LLAVA}, BLIP2 \citep{li2023blip2}, and InternVL \citep{chen2024internvl} commonly use pre-trained frozen vision models and train lightweight alignment layers to align visual features with existing linguistic embeddings present in (large) language models \citep{liu2024improvedbaselines}.
The frozen ViT version of CLIP is particularly often the basis for many current state-of-the-art VLMs (BLIP2, Molmo, InternVL, LLaVa, i.a.). 
If CLIP, as a backbone, does not consistently exhibit human-like cross-modal associations, it is difficult to imagine how additional alignment layers can capture human-like representations without extensive fine-tuning. 
Moreover, CLIP enables us to extend existing approaches with an interpretability-based methodology that cannot be applied to proprietary models.\footnote{The source data and code are available at \url{https://osf.io/gqsv6/}}

\paragraph{Linguistic inputs}
% Linguistic inputs
Models are probed through combining images and labels, the latter of which are pseudowords originating from four different sources. 
First, two sets of `original' labels are used, which have been traditionally used the most in human studies (\emph{bouba-kiki} and \emph{maluma-takete}), allowing for explicit comparison of human results with those of VLMs. 
Second, we borrow English adjectives from \citet{alper2024kiki} that are 10 synonyms of `sharp' and `curved'. 
These serve as a baseline informing us whether the models can, in principle, make correct cross-modal associations.
This differs from \citet{alper2024kiki}, who use adjectives to create a vector to calculate a geometric score.
Third, two-syllable labels were constructed following \citet{nielsen2013parsing}, using previously established cross-modal patterns in English. 
Sonorant consonants (M, N, L) and rounded vowels (OO, OH, AH) tend to match curved shapes, while plosive consonants (T, K, P) and non-rounded vowels (EE, AY, UH) align with jagged shapes.
Combining these yields 36 syllables (e.g., loo, nah, kee, puh), categorised into four types: sonorant-rounded (S-R), plosive-rounded (P-R), sonorant-non-rounded (S-NR), and plosive-non-rounded (P-NR). 
These were paired into two-syllable pseudowords, loosely replicating the task humans did in \citet{nielsen2013parsing}.
For analysis, we focus on `pure' pseudowords that are either fully round-associated (S-R-S-R) or fully sharp-associated (P-NR-P-NR), yielding 162 labels.
Fourth, VLMs are tested on pseudowords generated using the method from \citet{alper2024kiki}. 
Each pseudoword follows a three-syllable structure, where consonants (P, T, K, S, H, X, B, D, G, M, N, L) are combined with vowels (E, I, O, U, A).\footnote{The letter A is included per \citet{alper2024kiki} despite that it is typically not regarded to evoke cross-modal associations in humans.} 
Pseudowords repeat the first syllable at the end (e.g., `kitaki', `bodubo').
Only `pure' items—composed entirely of syllables from a single class—are used, excluding mixed forms like `kiduki'.

The linguistic inputs are comprised of a label of interest—which should induce certain preferences—and a prompt (`The label for this image is <label>') such that embedding the label happens at the sentence level and is closer to the models' natural objective. 
Importantly, for each image, we \textit{only} differ the pseudowords in question, so variation in probability must be a result of the pseudoword.
Provided that the pseudowords are generated anew, it is unlikely that the models encountered them during training.
Ten different prompts are used (see \autoref{appendix:linguisticinputs}) to ensure that the results are robust and not an artefact of peculiarities in the prompts.
These prompts are sourced from \citet{verhoef-etal-2024-kiki}, \citet{alper2024kiki}, or are newly created such that half of the labels appear as a noun, and the other half as an adjective.

% Visual inputs
\paragraph{Visual inputs}
The images fed to the models are either curved or jagged. 
Some of these images are sourced directly from previous works involving human participants \citep{kohler1929gestalt, kohler1947gestalt, maurer2006boubas, westbury2005soundsymbolism}, while others are specifically generated to test cross-modal associations in VLMs.
These were inspired by the method described in \citet{nielsen2013parsing} and already used by \citet{verhoef-etal-2024-kiki}. 
The generated images were created by randomly distributing points within a circle and then connecting them sequentially using curved lines for curved images and straight line segments for jagged images.
Hence, this method generated image pairs that subtly differ \textit{only} in features that are seemingly important for cross-modal effects. 
This differs from \citet{alper2024kiki}, who generated images using Stable Diffusion, leaving less experimental control and potentially distorting the understanding of which features cause the observed effects.
An overview of all image pairs is displayed in \autoref{appendix:visualinputs}.

\paragraph{Analyses}\label{analyses}
All our analyses use Bayesian Regression Models as implemented in the \textit{brms} package \citep{brms} in R \citep{R}. 
We fit models (using 4 chains of 4000 iterations and a warm-up of 2000) to predict the proportion of correct guesses given a \textit{Word\_type} with fixed effects for \textit{model}, \textit{prompt}, \textit{image pair}, or \textit{label pair}. 
The exact model formulas are displayed under each figure. 
The interpretation of our visualisation is straightforward: an effect is significant if the posterior means \textit{and} their credible intervals are above the chance level.

\section{Probing through probabilities}\label{exp1}
To assess the preferences of the ViT and ResNet versions, we extract probabilities for all possible labels (i.e., syllables and pseudowords) conditioned on an image.
For each image, we consider the label with the highest probability to be the model's preference (shown on the left in \autoref{fig:method-overview}). 
This is comparable to how \citet{nielsen2013parsing} tested for human preferences.
While using the probability data for all labels would have been possible, previous work demonstrated that using only the best-matching label yielded a small bouba-kiki effect, whereas using all probabilities did not \citep{verhoef-etal-2024-kiki}; as such, we use the most promising method.
Importantly, we follow \citet{cwiek2022acrossculture}, who rightfully treated human responses as bouba/kiki-congruent when participants matched bouba to a round shape, and crucially, also matched kiki to a spiky shape. 
In our work, this means that an image pair of a sharp \textit{and} round shape must be matched with a congruent label.

\begin{figure}
    \centering
    \includegraphics[width=\linewidth]{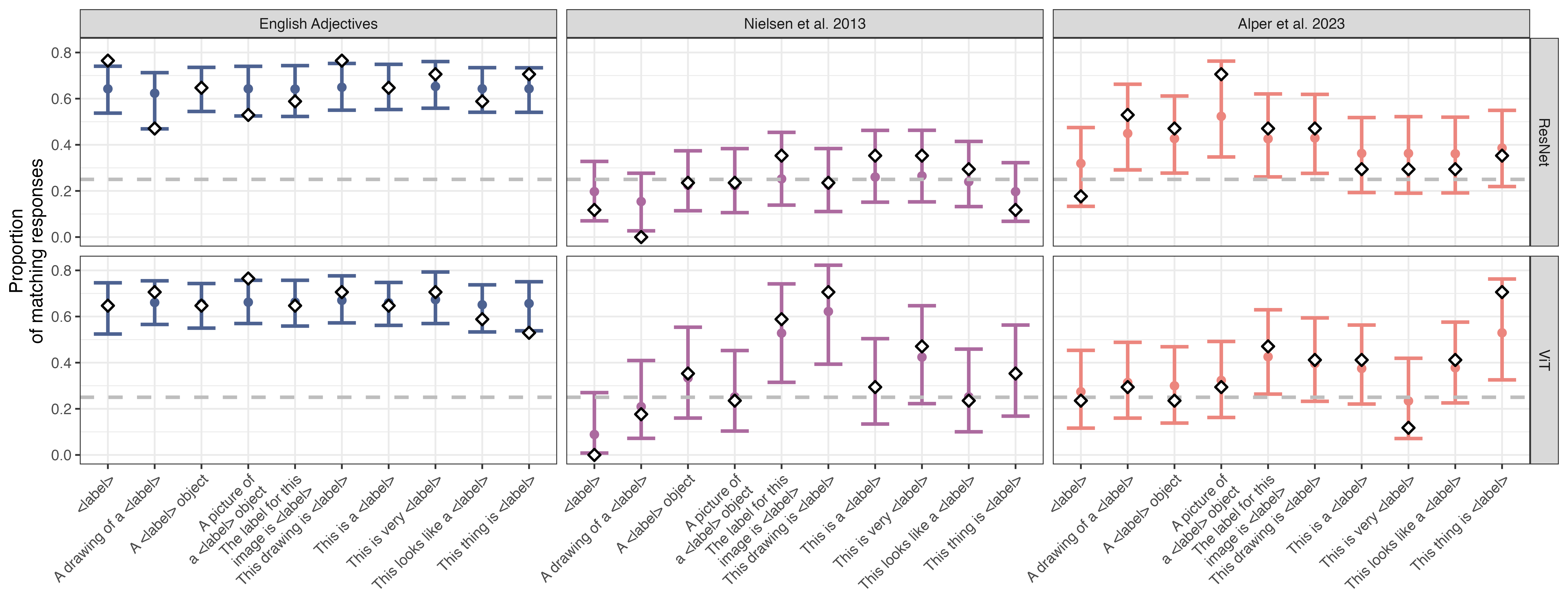}
    \caption{The proportion of congruent responses for matching both images of an image pair correctly ($Match = 1$). A result in which models consistently match images above chance (the grey dashed line) across prompts would suggest the presence of cross-modal associations. This is only the case for the English adjectives, which function as a baseline. Model: $Match \sim 1 + Word\_type + (1 + Word\_type | Prompt)$. Diamonds are descriptive means, and the dots are posterior means.}
    \label{fig:probability_winners}
\end{figure}

\paragraph{Results} 
We begin by considering the English adjectives, which serve as a baseline and do not inform us about the bouba-kiki effect. 
\autoref{fig:probability_winners} reveals that these are successfully matched to images with round or sharp features. 
This result is consistent across prompts and is also somewhat expected. 
Yet, a performance of roughly 80\% across prompts and models could also be considered low given the clear-cut distinctions between images and adjectives. 
The primary labels of interest (i.e., the pseudowords) can not be robustly and correctly matched to images with corresponding features by either model version. 
Despite some variability across prompts and models, with some combinations leading to above-chance performance (e.g., ViT, \citeauthor{nielsen2013parsing}'s pseudowords, and \textit{This drawing is <label>} or ResNet, \citeauthor{alper2024kiki}'s pseudowords, and \textit{A picture of a <label> object}), the descriptive means are mostly at chance level. 
Hence, we conclude that neither model displays clear cross-modal preferences. 
Interestingly, the individual modalities are, in principle, separable by both models (\ref{appendix:textual_embeddings} and \ref{appendix:visual_embeddings}), suggesting that the difficulty lies in combining information from multiple modalities.
This is also visible when looking into which labels are most commonly chosen, revealing that model predictions do not differ much even though they are presented with different images (\autoref{appendix:ratio_unique_labels}). 
For the experimental pseudowords, models seem to rely on a few frequently selected labels (the percentage of unique labels for ResNet and ViT is $\approx 35\%$). 
This strengthens our observation that no robust syllable-level cross-modal associations influence predictions.

\section{Beyond behavioural observations}\label{exp2}
In addition to using probability data, we are interested in knowing whether the predictions are wrong for the right reasons, i.e., do they fail to attend to curved or jagged regions when presented with our images?
As such, we further analyse the behaviour and preferences of models when associating an image with a label, using Grad-CAM, a technique from the model interpretability literature \citep{Selvaraju2019Grad-CAM}. 
Grad-CAM offers a visual explanation of a model's prediction by calculating the gradients of the target class score. 
This involves using the cosine similarity between the label and image embeddings in relation to the feature maps from the final convolutional layer or the last attention block. 
In the case of ViT, we applied Grad-CAM to the last attention layer, which retains spatial information via its attention heads.
We specifically focused on the attention from the [CLS] token to the image patches, similar to the approach in \cite{caron2021emergingproperties} and following public implementations of Grad-CAM \citep{zakka2021clipplayground, mamooler2021clipexplainability, Chefer_2021_ICCV}. 
% To identify which image regions contributed most to the decision, we computed the gradients of the similarity score between the image and text embeddings (which we used as the target for classification) with respect to the attention weights. 
Specifically, to identify which image regions contributed most to the decision, we computed the gradients of the target score with respect to the attention weights. 
These are averaged across heads, and we then extract the attention weights from the [CLS] token to the image patches by removing the [CLS] column. 
This is reshaped into a 2D spatial grid that acts as a feature map.
The values in these feature maps (typically displayed as a heatmap) can be interpreted as the importance or contribution of a specific region in the image to the model's prediction (see \autoref{appendix:gradcam} for some visual examples).

This technique enables us to very closely mimic the cross-cultural study conducted by \citet{cwiek2022acrossculture} in which human participants were shown the two classical bouba-kiki shapes and listened to the spoken words bouba or kiki. 
Hereafter, they selected which of the two shapes they thought corresponded to the word.
To simulate this experiment with VLMs, we concatenate all image pairs into a single image containing a curved and jagged shape (n=17). 
We then identify, for each label—which defines the expected target—which image region (left or right) receives the most attention from a model using Grad-CAM (see \autoref{fig:method-overview}). 
While the generated heatmaps visually indicate attention patterns, we quantify attention by summing the attention allocated to each curved and jagged part of the image.
Taking the total attention allocated to each part, rather than focusing on very specific regions, aligns with how humans perceive faces, objects, and words holistically \citep{Taubert2011TheRO, Zhao2016hollisticprocessing, wong2011wordshollistic}. 
Nevertheless, we also experimented with entropy and centroid-of-attention as quantifying measures, but none of these changed the results meaningfully (\autoref{appendix:gradcam}.
We compare the sum of importance values in the expected region for a given pseudoword within a particular category with the sum of importance values in the non-expected region for each image and text label. 
Based on this comparison, we compute the percentage of trials in which the model consistently focuses more on the expected region than on the non-expected region (as shown on the right in \autoref{fig:method-overview}).
The way images are concatenated is balanced such that the target appears eight or nine times on the left or right. 
Doing this eliminates a potential model bias toward objects on the left or right. 
Initial analyses across models, prompts, labels, and images revealed that both models are relatively consistent in their predictions, regardless of the target location, with ResNet being consistent for $77.0\%$ of the predictions and the ViT variant $73.5\%$ (\autoref{appendix:gradcam}).

\paragraph{Comparing VLMs with humans}
Using Grad-CAM, we compare the performance of both models to that of English-speaking participants as reported by \citet{cwiek2022acrossculture}. 
While they only investigated the well-known label pair bouba-kiki, another pair, maluma-takete, combined with the two images shown in \autoref{fig:method-overview}, was first described by Wolfgang K\"ohler \citep{kohler1929gestalt,kohler1947gestalt} and sparked wider interest in cross-modal associations.
We use only these two label pairs here, since the larger pseudoword set does not contain paired labels.
However, our image generation process, which only varies how points are connected, allows evaluating across all image pairs, beyond just the original images.
Specifically, we evaluate whether models attend more to the expected region (e.g., the sharp shape for `kiki' and `takete' or the round shape for `bouba' and `maluma').
A prediction is considered congruent only if both labels are correctly matched to their corresponding shapes. 
This setup closely mirrors human experiments and, for the bouba-kiki label pair, enables direct comparisons.

\begin{figure}
    \centering
    \includegraphics[width=\linewidth]{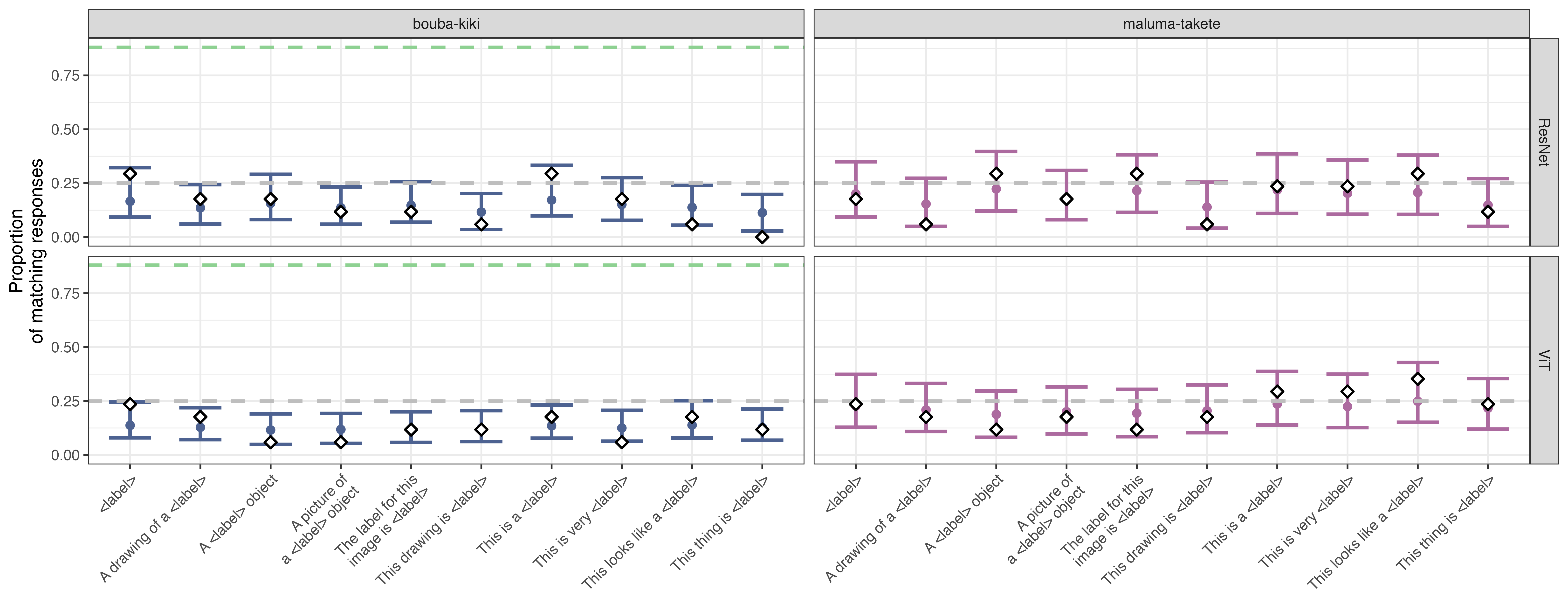}
    \caption{The proportion of correctly matched responses for both labels of a label pair ($Correct = 1$) given an image pair using Grad-CAM. The green line indicates human `performance' reported in \citep{cwiek2022acrossculture}. The grey line shows the chance level (25\%) as the model must map `bouba' \textit{and} `kiki' correctly. Model: $Correct \sim 1 + LabelPair + (1 + LabelPair | Prompt)$. Diamonds are descriptive means, and the dots are posterior means.}
    \label{fig:attention_boubakiki}
\end{figure}

Strikingly, neither CLIP model consistently maps both labels to their intended shapes at a human-like level (\autoref{fig:attention_boubakiki}); in fact, performance does not reliably exceed chance. 
These results contradict previous claims that vision–language models exhibit human-like cross-modal associations \citep{alper2024kiki}, and instead reinforce earlier findings showing no such effect.
If there is any situation in which an effect would be expected, it would be here since these classic word pairs are most dominantly present in the data.
Yet, these results show that merely learning \textit{about} an existing cross-modal effect from data distributions is different from having a mechanistic preference for cross-modal associations, which should not come as a surprise. 

\begin{figure}[b]
    \centering
    \includegraphics[width=\linewidth]{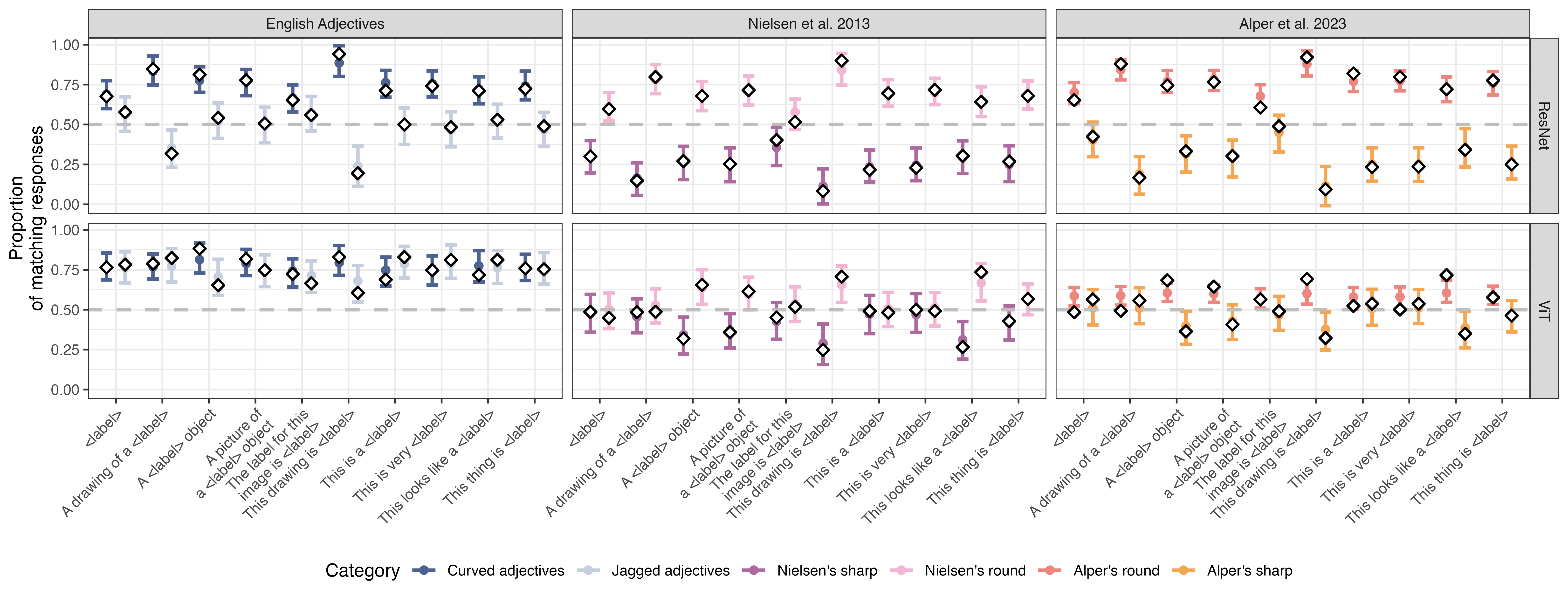}
    \caption{The proportion of correct matches (\textit{CorrectProportion}) given a word type and category (i.e., curved or sharp). A cross-modal association is indicated when a model consistently matches images above chance (grey line) for both categories—observed only with ViT and the English adjectives used for comparison. Model: $CorrectProportion \sim 1 + Word\_type + Category + (1 + Word\_type + Category | Prompt)$. The diamonds are descriptive averages, and the dots are posterior means.}
    \label{fig:attention}
\end{figure}

\paragraph{Analysing pseudowords}
Extending the analyses beyond the original word pairs, we now assess all pseudowords to examine whether CLIP displays a fundamental association between syllables and shapes. 
The proportion of pseudoword-label responses displayed in \autoref{fig:attention} reveals that there is again some variability across prompts, but model preferences are relatively consistent.
Using this method, the ResNet variant displays a general preference to attend to round shapes across all word types (note that not attending to a jagged shape if the label is of the jagged category means the model attends to the round shape).
Provided that this is also the case for the English adjectives, it is unsurprising that the pseudowords do not display a bouba-kiki effect.
The ViT variant is relatively consistent and can reliably distinguish shapes for English adjectives.
Yet, using experimental pseudowords collapses performance to chance levels, indicating the absence of human-like associations between shape features and syllables or characters. 
A qualitative inspection of several attention patterns provides additional evidence against the presence of cross-modal associations in CLIP.
This reveals that attention patterns primarily do not focus on sharp edges or round attributes of images, but instead mainly focus on the centres of shapes or background areas (\autoref{appendix:gradcam}). 
The latter is similar to earlier findings by \citet{darcet2024vision} who found that ViT networks create artefacts at low-informative background areas for computational purposes rather than describing visual information. 
Both observations are in contrast with what, at its core, is required for a bouba-kiki-like effect.

\section{Discussion and conclusion}
This work investigated whether vision-and-language models, specifically two versions of CLIP (ResNet and ViT), exhibit human-like patterns of cross-modal association, as reflected in the bouba-kiki effect. 
We aimed to rigorously evaluate the presence or absence of this phenomenon in model behaviour, using two methodological approaches closely aligned with human experiments and introducing the use of interpretability tools in this domain to probe internal representations.
Crucially, we posit that we can only speak of a consistent model preference when results across both methodologies are congruent with human-like preferences.
Our first experiment, which used image-label probabilities to gauge model preference, showed no clear cross-modal preference for either model.
The second experiment utilised Grad-CAM to simulate experimental conditions in human studies.
Here, the ResNet variant, which showed an overall preference for round shapes, did not demonstrate behaviour consistent enough to qualify as a bouba-kiki effect by human standards. 
The ViT-based version of CLIP, widely used as a foundational component in many current multimodal systems, proved capable of matching shapes and English adjectives, but failed to do so for pseudowords, which have even weaker alignment than ResNet.
% The ViT-based CLIP, widely used as a foundational component in many current multimodal systems, showed even weaker alignment.
Contrary to prior work suggesting VLMs may encode bouba-kiki-like associations \citep{alper2024kiki}, our findings reveal little to no evidence that CLIP models, under these experimental conditions, exhibit consistent human-like mappings between pseudowords and visual shape features.
This lack of alignment suggests that, despite their impressive performance across many downstream vision-language tasks, models like CLIP do not appear to represent cross-modal associations in a cognitively grounded manner internally. 
Together, this raises questions about how cross-modal grounding is encoded or inherited in many larger state-of-the-art architectures.

Our findings complicate earlier claims about the presence of sound-symbolic associations in model embeddings. 
While \citet{alper2024kiki} found strong evidence for a bouba-kiki effect using embedding similarity scores in CLIP and Stable Diffusion, \citet{Iida2024iconicityVLMs} used the same method with Japanese VLMs, but failed to pick up robust bouba-kiki-like mappings, even though Japanese is rich in sound-symbolism \citep{dingemanse2012advances}. 
Instead, the embedding similarity scores seem to reflect language-specific regularities rather than general cross-modal preferences. 
Our approach deliberately avoided this by mirroring psycholinguistic testing paradigms with carefully controlled pairs of novel images. 
Corroborating other recent studies \citep{verhoef-etal-2024-kiki, loakman-etal-2024-ears}, this empirically grounded probing method yielded results opposite to those of other studies, showing no evidence of a bouba-kiki effect. 
Whereas \citet{verhoef-etal-2024-kiki} report limited, though not consistent, evidence for a bouba-kiki effect, our work differs since it uses a more comprehensive set of pseudowords, amending their analyses and suggesting that there is no general effect.
We also ran the same tests using English adjectives, and in contrast to pseudowords, the models are able to make the expected mappings in that case. 
Furthermore, by employing Grad-CAM in the domain of cross-modal associations, we examined whether models visually attend to shape features that correlate with pseudoword form profiles.
This enabled us not only to assess whether the models exhibit human-like cross-modal associations, but also to determine whether they do so for the right reasons by probing the decision-making pathways within the models. 
They largely did not attend to the expected shape features, which presents further evidence for the claim that VLMs may lack human-like cross-modal representations.

% WHY are the results the way they are?
% Reflection
What causes this misalignment between humans and VLMs?
The broader implication is that current VLMs, even those trained on massive paired datasets, lack a key component of human-like multisensory understanding: grounded, flexible, and intuitive mappings between sensory modalities. 
This is perhaps not surprising given the lack of embodied interaction in their training and reliance on statistical co-occurrence rather than perceptual salience. 
Still, it can be argued that the co-occurrences in the training data are full of human-like preferences, including cross-modal associations. 
So while we may expect VLMs to pick up on these associations, their internal mechanisms inherently differ from those of humans and do not necessarily learn the same preferences from aligned image-text pairs.
% Given that the internal mechanisms of VLMs also inherently differ from those of humans, simply learning from aligned image-text pairs that are influenced by human-like preferences, including cross-modal associations, does not imply that they also learn the same preferences. 
Previous work on visual grounding \citep{jones-etal-2024-multimodal}, and shortcut learning, i.e., solving a test using unexpected non-human-like shortcuts \citep{schwartz-stanovsky-2022-limitations, mitchell2023debateunderstanding} argued similarly.
However, interestingly, we observed that the individual modalities are, in principle, separable.
This suggests that CLIP’s training objective (i.e., a sentence-level contrastive loss) and its method for aligning text and image modalities (i.e., cosine similarity) are the culprits.
Neither seems to specifically promote learning relations between visual features and phonetic elements in language.

Another straightforward difference between humans and VLMs is tokenisation. 
Unlike humans, who perceive words holistically \citep{wong2011wordshollistic}, tokenisation can distort word representations, reducing the potential for human-like associations with visual shapes.  
Inspection of the tokenised pseudowords, however, reveals that this does not occur frequently (`Bouba' becomes `bou' and `ba', `kiki' is one token, `sepise' becomes `sep', `ise', and `kaykuh' becomes `kay', `ku', `h'. See section \ref{appendix:tokenisation} for more examples.)
Most phonological structures are preserved that could match shape features. 
However, tokenisation may still split some pseudowords in ways that wouldn't evoke the expected cross-modal associations in humans (e.g., `H' in `OH' might elicit jagged rather than curved associations). 
In the cases where tokenisation breaks pseudowords into syllables that also show no bouba-kiki effects in humans, the set of alternative pseudowords is large enough to contain options that could invoke these associations. 
% If models had a genuine preference, they would prefer pseudowords tokenised as complete syllables over those split into non-syllables. 
% Our results show this is not the case.
If models had a genuine preference in the direction of a bouba-kiki effect, they would prefer those pseudowords that are tokenised as complete syllables, and retain the potential cross-modal association of interest, over those split into non-syllables in which the association may be corrupted. 
Our results show this is not the case.
As such, the potential value of developing models with shared preferences between humans and machines remains significant. 
Establishing alignment in human and machine understanding of (visual-auditory) form-meaning mappings and mutual understanding can enhance their interactions \citep{kouwenhoven2022emerging}, helping in creating AI systems that resemble how humans process and communicate meaning. 
A potentially rich line of future work includes systematically exploring how design choices in VLMs affect the emergence of cognitively meaningful representations and consistently incorporating cross-linguistic studies to account for cultural variation (as demonstrated by \citet{Iida2024iconicityVLMs}).

Our work has some notable limitations. 
First, we only test on CLIP versions. 
We deliberately chose to focus on the CLIP model instead of more capable proprietary models (GPT-4o and Gemini2-Flash, i.a.), since they are not sufficiently transparent and thus do not help advance our understanding of their internal mechanisms.
Nevertheless, it remains interesting to further unravel cross-modal associations in more contemporary models. 
Although they often use frozen ViT variants of CLIP, the alignment layer between CLIP's embedding and the language model's embedding may learn preferences that resemble those of humans.
% While the original bouba-kiki effect is rooted in sound symbolism, our work—and that of others \cite[e.g.,][]{alper2024kiki, loakman-etal-2024-ears, Iida2024iconicityVLMs}—relies on text, which may influence the outcomes. 
% Yet, the relation between orthographic shapes and sounds is not arbitrary either; it likely reflects human iconic strategies shaped over time \citep{turoman2017glyph}. 
% Exploring audio-visual models, as in \citet{tseng-etal-2024-measuring}, offers a promising direction for future research.
% Finally, we rely on the sum of intensities to select a shape in our Grad-CAM experiment. 
% While this aligns with how humans perceive objects and words \citep{Zhao2016hollisticprocessing, wong2011wordshollistic}, it is worth further examining, perhaps more qualitatively, more fine-grained Grad-CAM metrics.
While the original bouba-kiki effect is rooted in sound symbolism, our work—and that of others \cite[e.g.,][]{alper2024kiki, loakman-etal-2024-ears, Iida2024iconicityVLMs}—relies on text, which may influence the outcomes.
Yet, prior human experiments present pseudowords both acoustically and in written form \citep[e.g.,][]{nielsen2013parsing}, making it challenging to disentangle orthographic from auditory contributions.
Moreover, \citet{cuskley2017phonological} showed that auditory presentation cannot eliminate orthographic effects in literate participants, as characters are strongly associated with particular sounds.
These associations are non-arbitrary: writing systems often employ iconic strategies, in which characters representing rounded sounds (rounded vowels, sonorants) tend to have curved shapes \citep{koriat1977symbolic, turoman2017glyph}.
Given this tight coupling between sound and orthography, correspondences between vision and spoken words should be highly correlated with vision-text correspondences.
This makes it unlikely that audio-visual models would show more substantial bouba-kiki effects than text-based models.
Nonetheless, exploring audio-visual models \citep{tseng-etal-2024-measuring} remains a valuable direction for understanding how different modalities contribute to cross-modal associations.

% [in the interest of space perhaps we should remove the conclusion as a separate section and just end with a concluding sentence like the one below?]
Overall, our results reinforce the importance of interpretability and cognitively inspired evaluation when assessing model performance in cross-modal reasoning. If VLMs are to serve as truly intuitive agents in real-world human-machine interactions, they must not only succeed on benchmark datasets but also exhibit a more human-like understanding of abstract, grounded concepts.

% \section*{Acknowledgements}
% If we need to acknowledge someone this can be done here. 

\bibliographystyle{apalike}
\bibliography{bibliography}

\begin{thebibliography}{}

\bibitem[Abramova and Fern{\'a}ndez, 2016]{abramova-fernandez-2016-questioning}
Abramova, E. and Fern{\'a}ndez, R. (2016).
\newblock Questioning arbitrariness in language: a data-driven study of conventional iconicity.
\newblock In Knight, K., Nenkova, A., and Rambow, O., editors, {\em Proceedings of the 2016 Conference of the North {A}merican Chapter of the Association for Computational Linguistics: Human Language Technologies}, pages 343--352, San Diego, California. Association for Computational Linguistics.

\bibitem[Agrawal et~al., 2018]{agrawal2018dont-just-assume}
Agrawal, A., Batra, D., Parikh, D., and Kembhavi, A. (2018).
\newblock Don't just assume; look and answer: Overcoming priors for visual question answering.
\newblock In {\em Proceedings of the IEEE conference on computer vision and pattern recognition}, pages 4971--4980.

\bibitem[Allen et~al., 2025]{allen2025the}
Allen, K.~R., Dasgupta, I., Kosoy, E., and Lampinen, A.~K. (2025).
\newblock The in-context inductive biases of vision-language models differ across modalities.
\newblock In {\em Second Workshop on Representational Alignment at ICLR 2025}.

\bibitem[Alper and Averbuch-Elor, 2023]{alper2024kiki}
Alper, M. and Averbuch-Elor, H. (2023).
\newblock Kiki or bouba? sound symbolism in vision-and-language models.
\newblock In Oh, A., Naumann, T., Globerson, A., Saenko, K., Hardt, M., and Levine, S., editors, {\em Advances in Neural Information Processing Systems}, volume~36, pages 78347--78359. Curran Associates, Inc.

\bibitem[Blasi et~al., 2016]{blasi2016sound}
Blasi, D.~E., Wichmann, S., Hammarstr{\"o}m, H., Stadler, P.~F., and Christiansen, M.~H. (2016).
\newblock Sound--meaning association biases evidenced across thousands of languages.
\newblock {\em Proceedings of the National Academy of Sciences}, 113(39):10818--10823.

\bibitem[Bürkner, 2021]{brms}
Bürkner, P.-C. (2021).
\newblock Bayesian item response modeling in {R} with {brms} and {Stan}.
\newblock {\em Journal of Statistical Software}, 100(5):1--54.

\bibitem[Caron et~al., 2021]{caron2021emergingproperties}
Caron, M., Touvron, H., Misra, I., J\'egou, H., Mairal, J., Bojanowski, P., and Joulin, A. (2021).
\newblock Emerging properties in self-supervised vision transformers.
\newblock In {\em Proceedings of the IEEE/CVF International Conference on Computer Vision (ICCV)}, pages 9650--9660.

\bibitem[Chefer et~al., 2021]{Chefer_2021_ICCV}
Chefer, H., Gur, S., and Wolf, L. (2021).
\newblock Generic attention-model explainability for interpreting bi-modal and encoder-decoder transformers.
\newblock In {\em Proceedings of the IEEE/CVF International Conference on Computer Vision (ICCV)}, pages 397--406.

\bibitem[Chen et~al., 2024]{chen2024internvl}
Chen, Z., Wu, J., Wang, W., Su, W., Chen, G., Xing, S., Zhong, M., Zhang, Q., Zhu, X., Lu, L., Li, B., Luo, P., Lu, T., Qiao, Y., and Dai, J. (2024).
\newblock Internvl: Scaling up vision foundation models and aligning for generic visual-linguistic tasks.
\newblock In {\em Proceedings of the IEEE/CVF Conference on Computer Vision and Pattern Recognition (CVPR)}, pages 24185--24198.

\bibitem[Cuskley and Kirby, 2013]{cuskley2013syn}
Cuskley, C. and Kirby, S. (2013).
\newblock {Synesthesia, Cross-Modality, and Language Evolution}.
\newblock In {\em {Oxford Handbook of Synesthesia}}. Oxford University Press.

\bibitem[Cuskley et~al., 2017]{cuskley2017phonological}
Cuskley, C., Simner, J., and Kirby, S. (2017).
\newblock Phonological and orthographic influences in the bouba--kiki effect.
\newblock {\em Psychological research}, 81:119--130.

\bibitem[{\'C}wiek et~al., 2022]{cwiek2022acrossculture}
{\'C}wiek, A., Fuchs, S., Draxler, C., Asu, E.~L., Dediu, D., Hiovain, K., Kawahara, S., Koutalidis, S., Krifka, M., Lippus, P., Lupyan, G., Oh, G.~E., Paul, J., Petrone, C., Ridouane, R., Reiter, S., Sch{\"u}mchen, N., Szalontai, {\'A}., {\"U}nal-Logacev, {\"O}., Zeller, J., Perlman, M., and Winter, B. (2022).
\newblock The bouba/kiki effect is robust across cultures and writing systems.
\newblock {\em Philosophical Transactions of the Royal Society B: Biological Sciences}, 377(1841):20200390.

\bibitem[Darcet et~al., 2024]{darcet2024vision}
Darcet, T., Oquab, M., Mairal, J., and Bojanowski, P. (2024).
\newblock Vision transformers need registers.
\newblock In {\em The Twelfth International Conference on Learning Representations}.

\bibitem[de~Varda and Strapparava, 2022]{devarda2022Cross-Modal}
de~Varda, A.~G. and Strapparava, C. (2022).
\newblock A cross-modal and cross-lingual study of iconicity in language: Insights from deep learning.
\newblock {\em Cognitive Science}, 46(6):e13147.

\bibitem[Deitke et~al., 2024]{deitke2024molmopixmoopenweights}
Deitke, M., Clark, C., Lee, S., Tripathi, R., Yang, Y., Park, J.~S., Salehi, M., Muennighoff, N., Lo, K., Soldaini, L., Lu, J., Anderson, T., Bransom, E., Ehsani, K., Ngo, H., Chen, Y., Patel, A., Yatskar, M., Callison-Burch, C., Head, A., Hendrix, R., Bastani, F., VanderBilt, E., Lambert, N., Chou, Y., Chheda, A., Sparks, J., Skjonsberg, S., Schmitz, M., Sarnat, A., Bischoff, B., Walsh, P., Newell, C., Wolters, P., Gupta, T., Zeng, K.-H., Borchardt, J., Groeneveld, D., Nam, C., Lebrecht, S., Wittlif, C., Schoenick, C., Michel, O., Krishna, R., Weihs, L., Smith, N.~A., Hajishirzi, H., Girshick, R., Farhadi, A., and Kembhavi, A. (2024).
\newblock Molmo and pixmo: Open weights and open data for state-of-the-art vision-language models.

\bibitem[Demircan et~al., 2024]{demircan2024evaluating}
Demircan, C., Saanum, T., Pettini, L., Binz, M., Baczkowski, B.~M., Doeller, C.~F., Garvert, M.~M., and Schulz, E. (2024).
\newblock Evaluating alignment between humans and neural network representations in image-based learning tasks.
\newblock In {\em The Thirty-eighth Annual Conference on Neural Information Processing Systems}.

\bibitem[Dingemanse, 2012]{dingemanse2012advances}
Dingemanse, M. (2012).
\newblock Advances in the cross-linguistic study of ideophones.
\newblock {\em Language and Linguistics Compass}, 6(10):654--672.

\bibitem[Dingemanse et~al., 2016]{dingemanse2016sound}
Dingemanse, M., Schuerman, W., Reinisch, E., Tufvesson, S., and Mitterer, H. (2016).
\newblock What sound symbolism can and cannot do: Testing the iconicity of ideophones from five languages.
\newblock {\em Language}, 92(2):e117--e133.

\bibitem[Diwan et~al., 2022]{diwan-etal-2022-winoground}
Diwan, A., Berry, L., Choi, E., Harwath, D., and Mahowald, K. (2022).
\newblock Why is winoground hard? investigating failures in visuolinguistic compositionality.
\newblock In Goldberg, Y., Kozareva, Z., and Zhang, Y., editors, {\em Proceedings of the 2022 Conference on Empirical Methods in Natural Language Processing}, pages 2236--2250, Abu Dhabi, United Arab Emirates. Association for Computational Linguistics.

\bibitem[Dosovitskiy et~al., 2021]{dosovitskiy2021an}
Dosovitskiy, A., Beyer, L., Kolesnikov, A., Weissenborn, D., Zhai, X., Unterthiner, T., Dehghani, M., Minderer, M., Heigold, G., Gelly, S., Uszkoreit, J., and Houlsby, N. (2021).
\newblock An image is worth 16x16 words: Transformers for image recognition at scale.
\newblock In {\em International Conference on Learning Representations}.

\bibitem[Fort and Schwartz, 2022]{fort2022resolving}
Fort, M. and Schwartz, J.-L. (2022).
\newblock Resolving the bouba-kiki effect enigma by rooting iconic sound symbolism in physical properties of round and spiky objects.
\newblock {\em Scientific reports}, 12(1):19172.

\bibitem[Goyal et~al., 2017]{goyal2019vqa}
Goyal, Y., Khot, T., Summers-Stay, D., Batra, D., and Parikh, D. (2017).
\newblock Making the v in vqa matter: Elevating the role of image understanding in visual question answering.
\newblock In {\em Proceedings of the IEEE conference on computer vision and pattern recognition}, pages 6904--6913.

\bibitem[He et~al., 2016]{he2016deep}
He, K., Zhang, X., Ren, S., and Sun, J. (2016).
\newblock Deep residual learning for image recognition.
\newblock In {\em 2016 IEEE Conference on Computer Vision and Pattern Recognition (CVPR)}, pages 770--778.

\bibitem[Hockett, 1960]{hockett1960origin}
Hockett, C.~F. (1960).
\newblock The origin of speech.
\newblock {\em Scientific American}, 203:88--96.

\bibitem[Iida and Funakura, 2024]{Iida2024iconicityVLMs}
Iida, H. and Funakura, H. (2024).
\newblock Investigating iconicity in vision-and-language models: A case study of the bouba/kikieffect in japanese models.
\newblock In {\em Proceedings of the 46th Annual Conference of the Cognitive Science Society}, volume~46.

\bibitem[Imai and Kita, 2014]{imai2014sound}
Imai, M. and Kita, S. (2014).
\newblock The sound symbolism bootstrapping hypothesis for language acquisition and language evolution.
\newblock {\em Philosophical transactions of the Royal Society B: Biological sciences}, 369(1651):20130298.

\bibitem[Imai et~al., 2008]{imai2008sound}
Imai, M., Kita, S., Nagumo, M., and Okada, H. (2008).
\newblock Sound symbolism facilitates early verb learning.
\newblock {\em Cognition}, 109(1):54--65.

\bibitem[Jabri et~al., 2016]{jabri2016revisiting-vqa}
Jabri, A., Joulin, A., and Van Der~Maaten, L. (2016).
\newblock Revisiting visual question answering baselines.
\newblock In {\em European conference on computer vision}, pages 727--739. Springer.

\bibitem[Jones et~al., 2024]{jones-etal-2024-multimodal}
Jones, C.~R., Bergen, B., and Trott, S. (2024).
\newblock Do multimodal large language models and humans ground language similarly?
\newblock {\em Computational Linguistics}, 50(3):1415--1440.

\bibitem[Kamath et~al., 2023]{kamath-etal-2023-whats}
Kamath, A., Hessel, J., and Chang, K.-W. (2023).
\newblock What{'}s ``up'' with vision-language models? investigating their struggle with spatial reasoning.
\newblock In Bouamor, H., Pino, J., and Bali, K., editors, {\em Proceedings of the 2023 Conference on Empirical Methods in Natural Language Processing}, pages 9161--9175, Singapore. Association for Computational Linguistics.

\bibitem[K\"ohler, 1929]{kohler1929gestalt}
K\"ohler, W. (1929).
\newblock {\em Gestalt Psychology}.
\newblock New York: Horace Liveright.

\bibitem[K\"ohler, 1947]{kohler1947gestalt}
K\"ohler, W. (1947).
\newblock {\em Gestalt Psychology}.
\newblock (2nd ed.) New York: Horace Liveright.

\bibitem[Koriat and Levy, 1977]{koriat1977symbolic}
Koriat, A. and Levy, I. (1977).
\newblock The symbolic implications of vowels and of their orthographic representations in two natural languages.
\newblock {\em Journal of Psycholinguistic Research}, 6(2):93--103.

\bibitem[Kouwenhoven et~al., 2024]{kouwenhoven-etal-2024-curious}
Kouwenhoven, T., Peeperkorn, M., Van~Dijk, B., and Verhoef, T. (2024).
\newblock The curious case of representational alignment: Unravelling visio-linguistic tasks in emergent communication.
\newblock In Kuribayashi, T., Rambelli, G., Takmaz, E., Wicke, P., and Oseki, Y., editors, {\em Proceedings of the Workshop on Cognitive Modeling and Computational Linguistics}, pages 57--71, Bangkok, Thailand. Association for Computational Linguistics.

\bibitem[Kouwenhoven et~al., 2022]{kouwenhoven2022emerging}
Kouwenhoven, T., Verhoef, T., De~Kleijn, R., and Raaijmakers, S. (2022).
\newblock Emerging grounded shared vocabularies between human and machine, inspired by human language evolution.
\newblock {\em Frontiers in Artificial Intelligence}, 5:886349.

\bibitem[Lake et~al., 2017]{lake2017buildingmachines}
Lake, B.~M., Ullman, T.~D., Tenenbaum, J.~B., and Gershman, S.~J. (2017).
\newblock Building machines that learn and think like people.
\newblock {\em Behavioral and Brain Sciences}, 40:e253.

\bibitem[Li et~al., 2023]{li2023blip2}
Li, J., Li, D., Savarese, S., and Hoi, S. (2023).
\newblock Blip-2: bootstrapping language-image pre-training with frozen image encoders and large language models.
\newblock In {\em Proceedings of the 40th International Conference on Machine Learning}, ICML'23. JMLR.org.

\bibitem[Li et~al., 2022]{li2022blip}
Li, J., Li, D., Xiong, C., and Hoi, S. (2022).
\newblock Blip: Bootstrapping language-image pre-training for unified vision-language understanding and generation.
\newblock In {\em International conference on machine learning}, pages 12888--12900. PMLR.

\bibitem[Liu et~al., 2024]{liu2024improvedbaselines}
Liu, H., Li, C., Li, Y., and Lee, Y.~J. (2024).
\newblock Improved baselines with visual instruction tuning.
\newblock In {\em Proceedings of the IEEE/CVF Conference on Computer Vision and Pattern Recognition (CVPR)}, pages 26296--26306.

\bibitem[Liu et~al., 2023]{liu2023LLAVA}
Liu, H., Li, C., Wu, Q., and Lee, Y.~J. (2023).
\newblock Visual instruction tuning.
\newblock In Oh, A., Naumann, T., Globerson, A., Saenko, K., Hardt, M., and Levine, S., editors, {\em Advances in Neural Information Processing Systems}, volume~36, pages 34892--34916. Curran Associates, Inc.

\bibitem[Loakman et~al., 2024]{loakman-etal-2024-ears}
Loakman, T., Li, Y., and Lin, C. (2024).
\newblock With ears to see and eyes to hear: Sound symbolism experiments with multimodal large language models.
\newblock In Al-Onaizan, Y., Bansal, M., and Chen, Y.-N., editors, {\em Proceedings of the 2024 Conference on Empirical Methods in Natural Language Processing}, pages 2849--2867, Miami, Florida, USA. Association for Computational Linguistics.

\bibitem[Mamooler, 2021]{mamooler2021clipexplainability}
Mamooler, S. (2021).
\newblock Clip explainability.

\bibitem[Marklová et~al., 2025]{marklová2025iconicitylargelanguagemodels}
Marklová, A., Milička, J., Ryvkin, L., Ľudmila Lacková~Bennet, and Kormaníková, L. (2025).
\newblock Iconicity in large language models.

\bibitem[Maurer et~al., 2006a]{maurer2006boubas}
Maurer, D., Pathman, T., and Mondloch, C.~J. (2006a).
\newblock The shape of boubas: Sound--shape correspondences in toddlers and adults.
\newblock {\em Developmental science}, 9(3):316--322.

\bibitem[Maurer et~al., 2006b]{maurer2006shape}
Maurer, D., Pathman, T., and Mondloch, C.~J. (2006b).
\newblock The shape of boubas: Sound--shape correspondences in toddlers and adults.
\newblock {\em Developmental science}, 9(3):316--322.

\bibitem[Mitchell and Krakauer, 2023]{mitchell2023debateunderstanding}
Mitchell, M. and Krakauer, D.~C. (2023).
\newblock The debate over understanding in ai’s large language models.
\newblock {\em Proceedings of the National Academy of Sciences}, 120(13):e2215907120.

\bibitem[Nielsen and Rendall, 2011]{nielsen2011sound}
Nielsen, A. and Rendall, D. (2011).
\newblock The sound of round: evaluating the sound-symbolic role of consonants in the classic takete-maluma phenomenon.
\newblock {\em Canadian Journal of Experimental Psychology/Revue canadienne de psychologie exp{\'e}rimentale}, 65(2):115.

\bibitem[Nielsen and Rendall, 2012]{nielsen2012source}
Nielsen, A. and Rendall, D. (2012).
\newblock The source and magnitude of sound-symbolic biases in processing artificial word material and their implications for language learning and transmission.
\newblock {\em Language and cognition}, 4(2):115--125.

\bibitem[Nielsen and Rendall, 2013]{nielsen2013parsing}
Nielsen, A. and Rendall, D. (2013).
\newblock Parsing the role of consonants versus vowels in the classic takete-maluma phenomenon.
\newblock {\em Canadian Journal of Experimental Psychology/Revue canadienne de psychologie exp{\'e}rimentale}, 67(2):153.

\bibitem[Perniss et~al., 2010]{perniss2010iconicity}
Perniss, P., Thompson, R., and Vigliocco, G. (2010).
\newblock Iconicity as a general property of language: evidence from spoken and signed languages.
\newblock {\em Frontiers in Psychology}, 1(227).

\bibitem[Perry et~al., 2015]{perry2015iconicity}
Perry, L.~K., Perlman, M., and Lupyan, G. (2015).
\newblock Iconicity in english and spanish and its relation to lexical category and age of acquisition.
\newblock {\em PloS one}, 10(9):e0137147.

\bibitem[Perry et~al., 2018]{perry2018iconicity}
Perry, L.~K., Perlman, M., Winter, B., Massaro, D.~W., and Lupyan, G. (2018).
\newblock Iconicity in the speech of children and adults.
\newblock {\em Developmental Science}, 21(3):e12572.

\bibitem[Pimentel et~al., 2019]{pimentel-etal-2019-meaning}
Pimentel, T., McCarthy, A.~D., Blasi, D., Roark, B., and Cotterell, R. (2019).
\newblock Meaning to form: Measuring systematicity as information.
\newblock In Korhonen, A., Traum, D., and M{\`a}rquez, L., editors, {\em Proceedings of the 57th Annual Meeting of the Association for Computational Linguistics}, pages 1751--1764, Florence, Italy. Association for Computational Linguistics.

\bibitem[{R Core Team}, 2024]{R}
{R Core Team} (2024).
\newblock {\em R: A Language and Environment for Statistical Computing}.
\newblock R Foundation for Statistical Computing, Vienna, Austria.

\bibitem[Radford et~al., 2021]{radford2021learning}
Radford, A., Kim, J.~W., Hallacy, C., Ramesh, A., Goh, G., Agarwal, S., Sastry, G., Askell, A., Mishkin, P., Clark, J., et~al. (2021).
\newblock Learning transferable visual models from natural language supervision.
\newblock In {\em International conference on machine learning}, pages 8748--8763. PMLR.

\bibitem[Ramachandran and Hubbard, 2001]{ramachandran2001synaesthesia}
Ramachandran, V.~S. and Hubbard, E.~M. (2001).
\newblock Synaesthesia--a window into perception, thought and language.
\newblock {\em Journal of Consciousness Studies}, 8(12):3--34.

\bibitem[Schwartz and Stanovsky, 2022]{schwartz-stanovsky-2022-limitations}
Schwartz, R. and Stanovsky, G. (2022).
\newblock On the limitations of dataset balancing: The lost battle against spurious correlations.
\newblock In Carpuat, M., de~Marneffe, M.-C., and Meza~Ruiz, I.~V., editors, {\em Findings of the Association for Computational Linguistics: NAACL 2022}, pages 2182--2194, Seattle, United States. Association for Computational Linguistics.

\bibitem[Selvaraju et~al., 2020]{Selvaraju2019Grad-CAM}
Selvaraju, R.~R., Cogswell, M., Das, A., Vedantam, R., Parikh, D., and Batra, D. (2020).
\newblock Grad-cam: Visual explanations from deep networks via gradient-based localization.
\newblock {\em Int. J. Comput. Vision}, 128(2):336–359.

\bibitem[Shiono et~al., 2025]{shiono-etal-2025-evaluating}
Shiono, D., Brassard, A., Ishizuki, Y., and Suzuki, J. (2025).
\newblock Evaluating model alignment with human perception: A study on shitsukan in {LLM}s and {LVLM}s.
\newblock In Rambow, O., Wanner, L., Apidianaki, M., Al-Khalifa, H., Eugenio, B.~D., and Schockaert, S., editors, {\em Proceedings of the 31st International Conference on Computational Linguistics}, pages 11428--11444, Abu Dhabi, UAE. Association for Computational Linguistics.

\bibitem[Tamariz et~al., 2018]{tamariz2018interactive}
Tamariz, M., Roberts, S.~G., Mart{\'\i}nez, J.~I., and Santiago, J. (2018).
\newblock The interactive origin of iconicity.
\newblock {\em Cognitive Science}, 42(1):334--349.

\bibitem[Taubert et~al., 2011]{Taubert2011TheRO}
Taubert, J., Apthorp, D., Aagten-Murphy, D., and Alais, D. (2011).
\newblock The role of holistic processing in face perception: Evidence from the face inversion effect.
\newblock {\em Vision Research}, 51:1273--1278.

\bibitem[Thrush et~al., 2022]{Thrush2022Winoground}
Thrush, T., Jiang, R., Bartolo, M., Singh, A., Williams, A., Kiela, D., and Ross, C. (2022).
\newblock Winoground: Probing vision and language models for visio-linguistic compositionality.
\newblock In {\em Proceedings of the IEEE/CVF Conference on Computer Vision and Pattern Recognition (CVPR)}, pages 5238--5248.

\bibitem[Tseng et~al., 2024]{tseng-etal-2024-measuring}
Tseng, W.-C., Shih, Y.-J., Harwath, D., and Mooney, R. (2024).
\newblock Measuring sound symbolism in audio-visual models.
\newblock In {\em 2024 IEEE Spoken Language Technology Workshop (SLT)}, pages 1165--1172.

\bibitem[Turoman and Styles, 2017]{turoman2017glyph}
Turoman, N. and Styles, S.~J. (2017).
\newblock Glyph guessing for ‘oo’and ‘ee’: Spatial frequency information in sound symbolic matching for ancient and unfamiliar scripts.
\newblock {\em Royal Society open science}, 4(9):170882.

\bibitem[Verhoef et~al., 2016a]{verhoef2016iconicity}
Verhoef, T., Kirby, S., and De~Boer, B. (2016a).
\newblock Iconicity and the emergence of combinatorial structure in language.
\newblock {\em Cognitive science}, 40(8):1969--1994.

\bibitem[Verhoef et~al., 2015]{verhoef2015emergence}
Verhoef, T., Roberts, S.~G., and Dingemanse, M. (2015).
\newblock Emergence of systematic iconicity: Transmission, interaction and analogy.
\newblock In {\em Proceedings of the Annual Meeting of the Cognitive Science Society}, pages 2481--2486. Cognitive Science Society.

\bibitem[Verhoef et~al., 2024]{verhoef-etal-2024-kiki}
Verhoef, T., Shahrasbi, K., and Kouwenhoven, T. (2024).
\newblock What does kiki look like? cross-modal associations between speech sounds and visual shapes in vision-and-language models.
\newblock In Kuribayashi, T., Rambelli, G., Takmaz, E., Wicke, P., and Oseki, Y., editors, {\em Proceedings of the Workshop on Cognitive Modeling and Computational Linguistics}, pages 199--213, Bangkok, Thailand. Association for Computational Linguistics.

\bibitem[Verhoef et~al., 2016b]{verhoef2016cognitive}
Verhoef, T., Walker, E., and Marghetis, T. (2016b).
\newblock Cognitive biases and social coordination in the emergence of temporal language.
\newblock In {\em Proceedings of the Annual Meeting of the Cognitive Science Society}, volume~38, pages 2615--2620.

\bibitem[Westbury, 2005a]{westbury2005implicit}
Westbury, C. (2005a).
\newblock Implicit sound symbolism in lexical access: Evidence from an interference task.
\newblock {\em Brain and language}, 93(1):10--19.

\bibitem[Westbury, 2005b]{westbury2005soundsymbolism}
Westbury, C. (2005b).
\newblock Implicit sound symbolism in lexical access: Evidence from an interference task.
\newblock {\em Brain and language}, 93(1):10--19.

\bibitem[Wong et~al., 2011]{wong2011wordshollistic}
Wong, A. C.-N., Bukach, C.~M., Yuen, C., Yang, L., Leung, S., and Greenspon, E. (2011).
\newblock Holistic processing of words modulated by reading experience.
\newblock {\em PLOS ONE}, 6(6):1--7.

\bibitem[Zakka, 2021]{zakka2021clipplayground}
Zakka, K. (2021).
\newblock A playground for clip-like models.

\bibitem[Zhao et~al., 2016]{Zhao2016hollisticprocessing}
Zhao, M., Bülthoff, H.~H., and Bülthoff, I. (2016).
\newblock Beyond faces and expertise: Facelike holistic processing of nonface objects in the absence of expertise.
\newblock {\em Psychological Science}, 27(2):213--222.
\newblock PMID: 26674129.

\end{thebibliography}

%%%%%%%%%%%%%%%%%%%%%%%%%%%%%%%%%%%%%%%%%%%%%%%%%%%%%%%%%%%%

\newpage
\section*{NeurIPS Paper Checklist}

\begin{enumerate}

\item {\bf Claims}
    \item[] Question: Do the main claims made in the abstract and introduction accurately reflect the paper's contributions and scope?
    \item[] Answer: \answerYes{} % Replace by \answerYes{}, \answerNo{}, or \answerNA{}.
    \item[] Justification: \autoref{exp1} clearly shows that both models perform below chance level. \autoref{exp2} directly compares computational results to work by \citep{cwiek2022acrossculture}.
    \item[] Guidelines:
    \begin{itemize}
        \item The answer NA means that the abstract and introduction do not include the claims made in the paper.
        \item The abstract and/or introduction should clearly state the claims made, including the contributions made in the paper and important assumptions and limitations. A No or NA answer to this question will not be perceived well by the reviewers. 
        \item The claims made should match theoretical and experimental results, and reflect how much the results can be expected to generalize to other settings. 
        \item It is fine to include aspirational goals as motivation as long as it is clear that these goals are not attained by the paper. 
    \end{itemize}

\item {\bf Limitations}
    \item[] Question: Does the paper discuss the limitations of the work performed by the authors?
    \item[] Answer: \answerYes{} % Replace by \answerYes{}, \answerNo{}, or \answerNA{}.
    \item[] Justification: The final section includes some limitations to our methodology, and reliance on text-based models instead of audio-visual models. We, moreover, discuss for each experiment how consistent the models' predictions are, regardless of whether they make human-like associations.
    \item[] Guidelines:
    \begin{itemize}
        \item The answer NA means that the paper has no limitation while the answer No means that the paper has limitations, but those are not discussed in the paper. 
        \item The authors are encouraged to create a separate "Limitations" section in their paper.
        \item The paper should point out any strong assumptions and how robust the results are to violations of these assumptions (e.g., independence assumptions, noiseless settings, model well-specification, asymptotic approximations only holding locally). The authors should reflect on how these assumptions might be violated in practice and what the implications would be.
        \item The authors should reflect on the scope of the claims made, e.g., if the approach was only tested on a few datasets or with a few runs. In general, empirical results often depend on implicit assumptions, which should be articulated.
        \item The authors should reflect on the factors that influence the performance of the approach. For example, a facial recognition algorithm may perform poorly when image resolution is low or images are taken in low lighting. Or a speech-to-text system might not be used reliably to provide closed captions for online lectures because it fails to handle technical jargon.
        \item The authors should discuss the computational efficiency of the proposed algorithms and how they scale with dataset size.
        \item If applicable, the authors should discuss possible limitations of their approach to address problems of privacy and fairness.
        \item While the authors might fear that complete honesty about limitations might be used by reviewers as grounds for rejection, a worse outcome might be that reviewers discover limitations that aren't acknowledged in the paper. The authors should use their best judgment and recognize that individual actions in favor of transparency play an important role in developing norms that preserve the integrity of the community. Reviewers will be specifically instructed to not penalize honesty concerning limitations.
    \end{itemize}

\item {\bf Theory Assumptions and Proofs}
    \item[] Question: For each theoretical result, does the paper provide the full set of assumptions and a complete (and correct) proof?
    \item[] Answer: \answerNA{}. % Replace by \answerYes{}, \answerNo{}, or \answerNA{}.
    \item[] Justification: Our work does not focus on making a theoretical contribution. 
    \item[] Guidelines:
    \begin{itemize}
        \item The answer NA means that the paper does not include theoretical results. 
        \item All the theorems, formulas, and proofs in the paper should be numbered and cross-referenced.
        \item All assumptions should be clearly stated or referenced in the statement of any theorems.
        \item The proofs can either appear in the main paper or the supplemental material, but if they appear in the supplemental material, the authors are encouraged to provide a short proof sketch to provide intuition. 
        \item Inversely, any informal proof provided in the core of the paper should be complemented by formal proofs provided in appendix or supplemental material.
        \item Theorems and Lemmas that the proof relies upon should be properly referenced. 
    \end{itemize}

    \item {\bf Experimental Result Reproducibility}
    \item[] Question: Does the paper fully disclose all the information needed to reproduce the main experimental results of the paper to the extent that it affects the main claims and/or conclusions of the paper (regardless of whether the code and data are provided or not)?
    \item[] Answer: \answerYes{} % Replace by \answerYes{}, \answerNo{}, or \answerNA{}.
    \item[] Justification: The method section (\autoref{sec:methods}) clearly describes how our linguistic inputs are made, or where they originate from. Idem for the images used. The Bayesian Regression Models used are also clearly specified in each analyses including the parameters used to fit them. 
    \item[] Guidelines:
    \begin{itemize}
        \item The answer NA means that the paper does not include experiments.
        \item If the paper includes experiments, a No answer to this question will not be perceived well by the reviewers: Making the paper reproducible is important, regardless of whether the code and data are provided or not.
        \item If the contribution is a dataset and/or model, the authors should describe the steps taken to make their results reproducible or verifiable. 
        \item Depending on the contribution, reproducibility can be accomplished in various ways. For example, if the contribution is a novel architecture, describing the architecture fully might suffice, or if the contribution is a specific model and empirical evaluation, it may be necessary to either make it possible for others to replicate the model with the same dataset, or provide access to the model. In general. releasing code and data is often one good way to accomplish this, but reproducibility can also be provided via detailed instructions for how to replicate the results, access to a hosted model (e.g., in the case of a large language model), releasing of a model checkpoint, or other means that are appropriate to the research performed.
        \item While NeurIPS does not require releasing code, the conference does require all submissions to provide some reasonable avenue for reproducibility, which may depend on the nature of the contribution. For example
        \begin{enumerate}
            \item If the contribution is primarily a new algorithm, the paper should make it clear how to reproduce that algorithm.
            \item If the contribution is primarily a new model architecture, the paper should describe the architecture clearly and fully.
            \item If the contribution is a new model (e.g., a large language model), then there should either be a way to access this model for reproducing the results or a way to reproduce the model (e.g., with an open-source dataset or instructions for how to construct the dataset).
            \item We recognize that reproducibility may be tricky in some cases, in which case authors are welcome to describe the particular way they provide for reproducibility. In the case of closed-source models, it may be that access to the model is limited in some way (e.g., to registered users), but it should be possible for other researchers to have some path to reproducing or verifying the results.
        \end{enumerate}
    \end{itemize}

\item {\bf Open access to data and code}
    \item[] Question: Does the paper provide open access to the data and code, with sufficient instructions to faithfully reproduce the main experimental results, as described in supplemental material?
    \item[] Answer: \answerYes{} % Replace by \answerYes{}, \answerNo{}, or \answerNA{}.
    \item[] Justification: Our source code and the data will be published on OSF upon publication. 
    \item[] Guidelines:
    \begin{itemize}
        \item The answer NA means that paper does not include experiments requiring code.
        \item Please see the NeurIPS code and data submission guidelines (\url{https://nips.cc/public/guides/CodeSubmissionPolicy}) for more details.
        \item While we encourage the release of code and data, we understand that this might not be possible, so “No” is an acceptable answer. Papers cannot be rejected simply for not including code, unless this is central to the contribution (e.g., for a new open-source benchmark).
        \item The instructions should contain the exact command and environment needed to run to reproduce the results. See the NeurIPS code and data submission guidelines (\url{https://nips.cc/public/guides/CodeSubmissionPolicy}) for more details.
        \item The authors should provide instructions on data access and preparation, including how to access the raw data, preprocessed data, intermediate data, and generated data, etc.
        \item The authors should provide scripts to reproduce all experimental results for the new proposed method and baselines. If only a subset of experiments are reproducible, they should state which ones are omitted from the script and why.
        \item At submission time, to preserve anonymity, the authors should release anonymized versions (if applicable).
        \item Providing as much information as possible in supplemental material (appended to the paper) is recommended, but including URLs to data and code is permitted.
    \end{itemize}

\item {\bf Experimental Setting/Details}
    \item[] Question: Does the paper specify all the training and test details (e.g., data splits, hyperparameters, how they were chosen, type of optimizer, etc.) necessary to understand the results?
    \item[] Answer: \answerYes{} % Replace by \answerYes{}, \answerNo{}, or \answerNA{}.
    \item[] Justification: Similar to the result reproducibility, our methodology is clearly described in \autoref{sec:methods}. The methods to create the images and labels are also described in detail. The tests used to make our contributions are also clearly specified underneath each figure. Moreover, the source code will be available upon publication, exposing any other experimental details.
    \item[] Guidelines:
    \begin{itemize}
        \item The answer NA means that the paper does not include experiments.
        \item The experimental setting should be presented in the core of the paper to a level of detail that is necessary to appreciate the results and make sense of them.
        \item The full details can be provided either with the code, in appendix, or as supplemental material.
    \end{itemize}

\item {\bf Experiment Statistical Significance}
    \item[] Question: Does the paper report error bars suitably and correctly defined or other appropriate information about the statistical significance of the experiments?
    \item[] Answer: \answerYes{} % Replace by \answerYes{}, \answerNo{}, or \answerNA{}.
    \item[] Justification: We provide credible intervals in each figure and explain how our analyses should be interpreted (\autoref{analyses}). Moreover, we explicitly state the models fitted and the parameters used to do the fitting. 
    \item[] Guidelines:
    \begin{itemize}
        \item The answer NA means that the paper does not include experiments.
        \item The authors should answer "Yes" if the results are accompanied by error bars, confidence intervals, or statistical significance tests, at least for the experiments that support the main claims of the paper.
        \item The factors of variability that the error bars are capturing should be clearly stated (for example, train/test split, initialization, random drawing of some parameter, or overall run with given experimental conditions).
        \item The method for calculating the error bars should be explained (closed form formula, call to a library function, bootstrap, etc.)
        \item The assumptions made should be given (e.g., Normally distributed errors).
        \item It should be clear whether the error bar is the standard deviation or the standard error of the mean.
        \item It is OK to report 1-sigma error bars, but one should state it. The authors should preferably report a 2-sigma error bar than state that they have a 96\% CI, if the hypothesis of Normality of errors is not verified.
        \item For asymmetric distributions, the authors should be careful not to show in tables or figures symmetric error bars that would yield results that are out of range (e.g. negative error rates).
        \item If error bars are reported in tables or plots, The authors should explain in the text how they were calculated and reference the corresponding figures or tables in the text.
    \end{itemize}

\item {\bf Experiments Compute Resources}
    \item[] Question: For each experiment, does the paper provide sufficient information on the computer resources (type of compute workers, memory, time of execution) needed to reproduce the experiments?
    \item[] Answer: \answerNo{} % Replace by \answerYes{}, \answerNo{}, or \answerNA{}.
    \item[] Justification: Our experiments do not need heavy computing since we do not train new models or fine-tune anything. All the code and analyses ran on a single MacBook Air M1. 
    \item[] Guidelines:
    \begin{itemize}
        \item The answer NA means that the paper does not include experiments.
        \item The paper should indicate the type of compute workers CPU or GPU, internal cluster, or cloud provider, including relevant memory and storage.
        \item The paper should provide the amount of compute required for each of the individual experimental runs as well as estimate the total compute. 
        \item The paper should disclose whether the full research project required more compute than the experiments reported in the paper (e.g., preliminary or failed experiments that didn't make it into the paper). 
    \end{itemize}
    
\item {\bf Code Of Ethics}
    \item[] Question: Does the research conducted in the paper conform, in every respect, with the NeurIPS Code of Ethics \url{https://neurips.cc/public/EthicsGuidelines}?
    \item[] Answer: \answerYes{} % Replace by \answerYes{}, \answerNo{}, or \answerNA{}.
    \item[] Justification: We do not violate the code of ethics. 
    \item[] Guidelines:
    \begin{itemize}
        \item The answer NA means that the authors have not reviewed the NeurIPS Code of Ethics.
        \item If the authors answer No, they should explain the special circumstances that require a deviation from the Code of Ethics.
        \item The authors should make sure to preserve anonymity (e.g., if there is a special consideration due to laws or regulations in their jurisdiction).
    \end{itemize}

\item {\bf Broader Impacts}
    \item[] Question: Does the paper discuss both potential positive societal impacts and negative societal impacts of the work performed?
    \item[] Answer: \answerNo{} % Replace by \answerYes{}, \answerNo{}, or \answerNA{}.
    \item[] Justification: Our work does not discuss societal impact, as it primarily focuses on whether current CLIP models (already widely available) align with human cross-modal preferences. We do not develop a new model that may be used by society, deal with biased data that may result in discrimination, or any related issue. As such, we do not deem it necessary to discuss societal impact. 
    \item[] Guidelines:
    \begin{itemize}
        \item The answer NA means that there is no societal impact of the work performed.
        \item If the authors answer NA or No, they should explain why their work has no societal impact or why the paper does not address societal impact.
        \item Examples of negative societal impacts include potential malicious or unintended uses (e.g., disinformation, generating fake profiles, surveillance), fairness considerations (e.g., deployment of technologies that could make decisions that unfairly impact specific groups), privacy considerations, and security considerations.
        \item The conference expects that many papers will be foundational research and not tied to particular applications, let alone deployments. However, if there is a direct path to any negative applications, the authors should point it out. For example, it is legitimate to point out that an improvement in the quality of generative models could be used to generate deepfakes for disinformation. On the other hand, it is not needed to point out that a generic algorithm for optimizing neural networks could enable people to train models that generate Deepfakes faster.
        \item The authors should consider possible harms that could arise when the technology is being used as intended and functioning correctly, harms that could arise when the technology is being used as intended but gives incorrect results, and harms following from (intentional or unintentional) misuse of the technology.
        \item If there are negative societal impacts, the authors could also discuss possible mitigation strategies (e.g., gated release of models, providing defenses in addition to attacks, mechanisms for monitoring misuse, mechanisms to monitor how a system learns from feedback over time, improving the efficiency and accessibility of ML).
    \end{itemize}
    
\item {\bf Safeguards}
    \item[] Question: Does the paper describe safeguards that have been put in place for responsible release of data or models that have a high risk for misuse (e.g., pretrained language models, image generators, or scraped datasets)?
    \item[] Answer: \answerNA{} % Replace by \answerYes{}, \answerNo{}, or \answerNA{}.
    \item[] Justification: There is no need for safeguards since we only evaluate already available models. 
    \item[] Guidelines:
    \begin{itemize}
        \item The answer NA means that the paper poses no such risks.
        \item Released models that have a high risk for misuse or dual-use should be released with necessary safeguards to allow for controlled use of the model, for example by requiring that users adhere to usage guidelines or restrictions to access the model or implementing safety filters. 
        \item Datasets that have been scraped from the Internet could pose safety risks. The authors should describe how they avoided releasing unsafe images.
        \item We recognize that providing effective safeguards is challenging, and many papers do not require this, but we encourage authors to take this into account and make a best faith effort.
    \end{itemize}

\item {\bf Licenses for existing assets}
    \item[] Question: Are the creators or original owners of assets (e.g., code, data, models), used in the paper, properly credited and are the license and terms of use explicitly mentioned and properly respected?
    \item[] Answer: \answerYes{} % Replace by \answerYes{}, \answerNo{}, or \answerNA{}.
    \item[] Justification: We cite the code-base we used for Grad-CAM (\autoref{exp2}), and properly mention the sources of the images and labels that we did not generate ourselves (\autoref{sec:methods}). 
    \item[] Guidelines:
    \begin{itemize}
        \item The answer NA means that the paper does not use existing assets.
        \item The authors should cite the original paper that produced the code package or dataset.
        \item The authors should state which version of the asset is used and, if possible, include a URL.
        \item The name of the license (e.g., CC-BY 4.0) should be included for each asset.
        \item For scraped data from a particular source (e.g., website), the copyright and terms of service of that source should be provided.
        \item If assets are released, the license, copyright information, and terms of use in the package should be provided. For popular datasets, \url{paperswithcode.com/datasets} has curated licenses for some datasets. Their licensing guide can help determine the license of a dataset.
        \item For existing datasets that are re-packaged, both the original license and the license of the derived asset (if it has changed) should be provided.
        \item If this information is not available online, the authors are encouraged to reach out to the asset's creators.
    \end{itemize}

\item {\bf New Assets}
    \item[] Question: Are new assets introduced in the paper well documented and is the documentation provided alongside the assets?
    \item[] Answer: \answerNA{} % Replace by \answerYes{}, \answerNo{}, or \answerNA{}.
    \item[] Justification: We do not release new assets.
    \item[] Guidelines:
    \begin{itemize}
        \item The answer NA means that the paper does not release new assets.
        \item Researchers should communicate the details of the dataset/code/model as part of their submissions via structured templates. This includes details about training, license, limitations, etc. 
        \item The paper should discuss whether and how consent was obtained from people whose asset is used.
        \item At submission time, remember to anonymize your assets (if applicable). You can either create an anonymized URL or include an anonymized zip file.
    \end{itemize}

\item {\bf Crowdsourcing and Research with Human Subjects}
    \item[] Question: For crowdsourcing experiments and research with human subjects, does the paper include the full text of instructions given to participants and screenshots, if applicable, as well as details about compensation (if any)? 
    \item[] Answer: \answerNA{} % Replace by \answerYes{}, \answerNo{}, or \answerNA{}.
    \item[] Justification: We do not use human subjects but base our comparisons to human performance on other work. 
    \item[] Guidelines:
    \begin{itemize}
        \item The answer NA means that the paper does not involve crowdsourcing nor research with human subjects.
        \item Including this information in the supplemental material is fine, but if the main contribution of the paper involves human subjects, then as much detail as possible should be included in the main paper. 
        \item According to the NeurIPS Code of Ethics, workers involved in data collection, curation, or other labor should be paid at least the minimum wage in the country of the data collector. 
    \end{itemize}

\item {\bf Institutional Review Board (IRB) Approvals or Equivalent for Research with Human Subjects}
    \item[] Question: Does the paper describe potential risks incurred by study participants, whether such risks were disclosed to the subjects, and whether Institutional Review Board (IRB) approvals (or an equivalent approval/review based on the requirements of your country or institution) were obtained?
    \item[] Answer: \answerNA{} % Replace by \answerYes{}, \answerNo{}, or \answerNA{}.
    \item[] Justification: Our work did not involve crowdsourcing nor research with human subjects. 
    \item[] Guidelines:
    \begin{itemize}
        \item The answer NA means that the paper does not involve crowdsourcing nor research with human subjects.
        \item Depending on the country in which research is conducted, IRB approval (or equivalent) may be required for any human subjects research. If you obtained IRB approval, you should clearly state this in the paper. 
        \item We recognize that the procedures for this may vary significantly between institutions and locations, and we expect authors to adhere to the NeurIPS Code of Ethics and the guidelines for their institution. 
        \item For initial submissions, do not include any information that would break anonymity (if applicable), such as the institution conducting the review.
    \end{itemize}

\end{enumerate}

%%%%%%%%%%%%%%%%%%%%%%%%%%%%%%%%%%%%%%%%%%%%%%%%%%%%%%%%%%%%%%%%%%%
% Checklist

\newpage

\appendix

\section{Linguistic inputs}\label{appendix:linguisticinputs}
The experimental inputs are tested across ten different prompts to assess robustness. 
They originate from earlier investigations \citep{alper2024kiki, verhoef-etal-2024-kiki} and are extended with additional prompts such that the \textit{label} occurs equally frequently as a noun and adjective (\autoref{appendix:tab:prompts}).

\begin{table}[h]
    \centering
    \begin{tabular}{lcr} \toprule
         Prompt                                         & Word type & Origin                        \\ \midrule
         \textit{The label for this image is <label>}   & Noun      & \cite{verhoef-etal-2024-kiki} \\
         \textit{This is a <label>}                     & Noun      & new                           \\
         \textit{A drawing of a <label>}                & Noun      & new                           \\
         \textit{This drawing is <label>}               & Adjective & new                           \\
         \textit{This thing is <label>}                 & Adjective & \cite{alper2024kiki}          \\
         \textit{A <label> object}                      & Adjective & \cite{alper2024kiki}          \\
         \textit{A picture of a <label> object}         & Adjective & \cite{alper2024kiki}          \\
         \textit{<label>}                               & Noun      & \cite{alper2024kiki}          \\
         \textit{This looks like a <label>}             & Noun      & new                           \\
         \textit{This is very <label>}                  & Adjective & new                           \\ \bottomrule
    \end{tabular}
    \caption{Different linguistic inputs are used to test our models for cross-modal associations.}
    \label{appendix:tab:prompts}
\end{table}

\subsection{Textual embeddings}\label{appendix:textual_embeddings}
\begin{figure}[h]
    \centering
    \includegraphics[width=\linewidth]{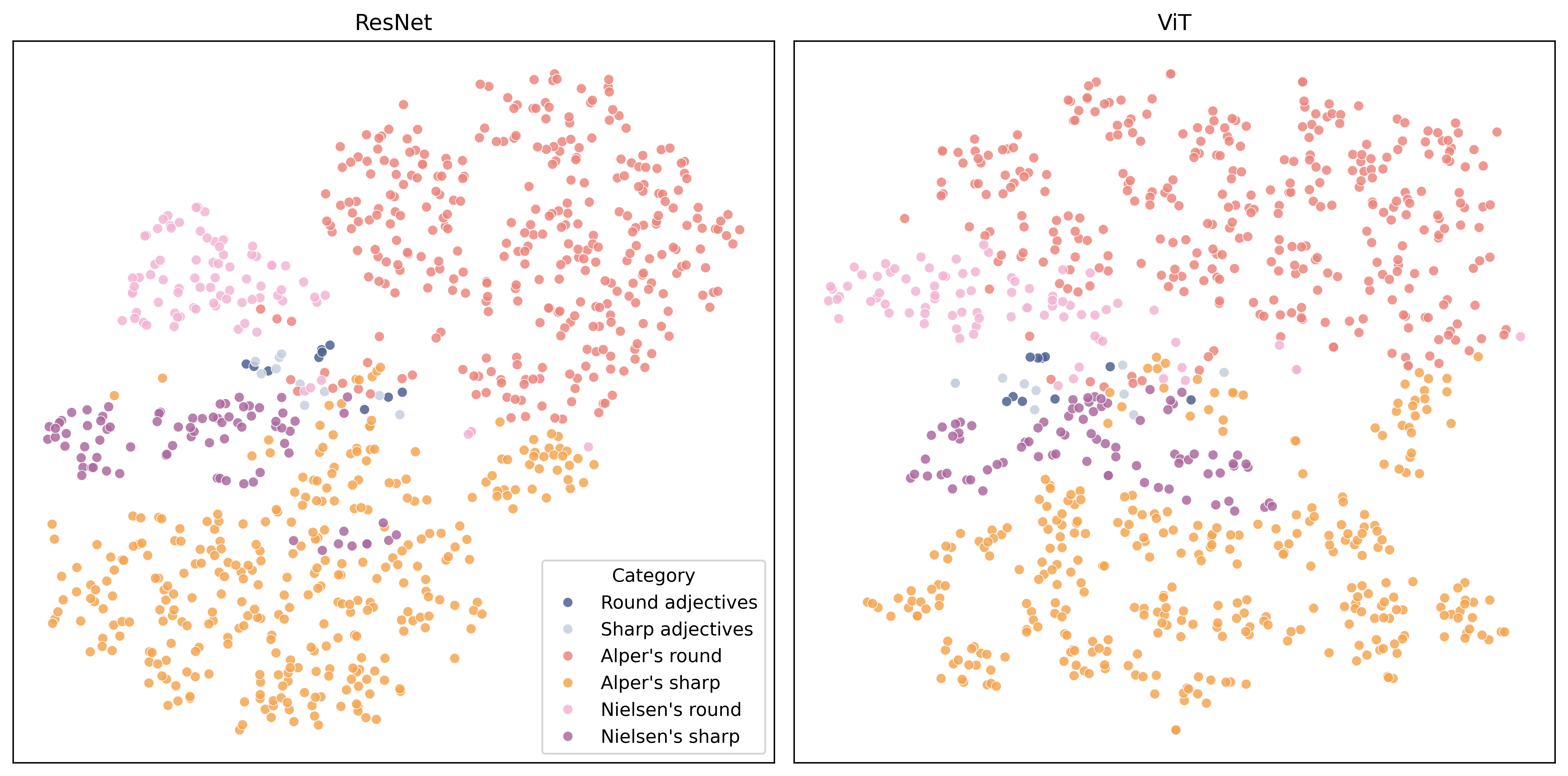}
    \caption{t-SNE plot showing how the language models of different CLIP variants interpret labels from different categories. The colour shades indicate which word type a label in a category belongs to. In order to correctly match labels to images with shape-specific features, a model must be able to discriminate word types between labels of the same category. This is clearly possible. This plot shows the embeddings for the prompt: \textit{The label for this image is <label>}. Different plots result in similar distributions.}
    \label{appendix:fig:tsne-label}
\end{figure}

To make correct associations, the models must, at a minimum, be able to disambiguate labels (real and non-words) from each other.
If they fail to do so, it is presumably also impossible to link certain words and their linguistic features to shape-specific features.
\autoref{appendix:fig:tsne-label} displays a t-SNE visualisation of the models' embeddings and reveals that they, in principle, should be able to disambiguate labels from different categories within a word type.

\subsection{Tokenisation}\label{appendix:tokenisation}
To determine whether byte-pair encodings or our pseudowords break phonological structures, we qualitatively assess a set of tokenised examples. 
These pseudowords are parsed as elements of the sentence prompts, but we only focus on the pseudowords themselves. 
The list below reveals that most of the target words are tokenised in a way that preserves at least some phonological structures that could be matched to shape features.

\begin{itemize}
    \item[-] Bouba → `bou' and `ba'
    \item[-] Kiki → `kiki'
    \item[-] Takete → `take', `te'
    \item[-] Maluma → `mal', `uma'
    \item[-] Xehaxe → `xe', `ha', `xe'
    \item[-] Lohlah → `loh', `lah'
    \item[-] Sepise → `sep', `ise'
    \item[-] Kaykuh → `kay', `ku', `h'
    \item[-] Loomoh → `loom', `oh'
    \item[-] Mohmah → `moh', `mah'
\end{itemize}

\section{Visual inputs} \label{appendix:visualinputs}
\begin{figure}
   \includegraphics[width=0.24\linewidth]{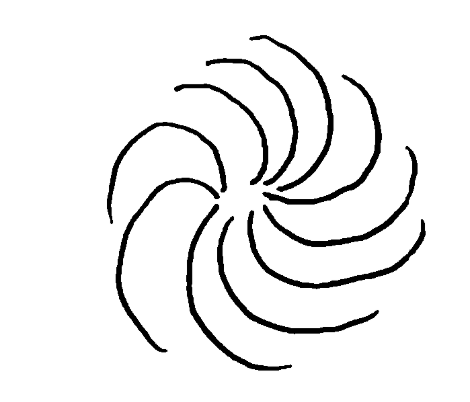}
   \includegraphics[width=0.24\linewidth]{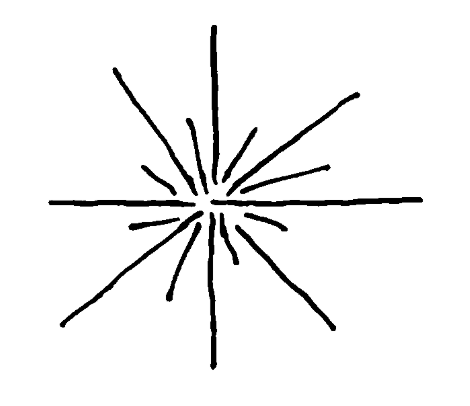}
   \includegraphics[width=0.24\linewidth]{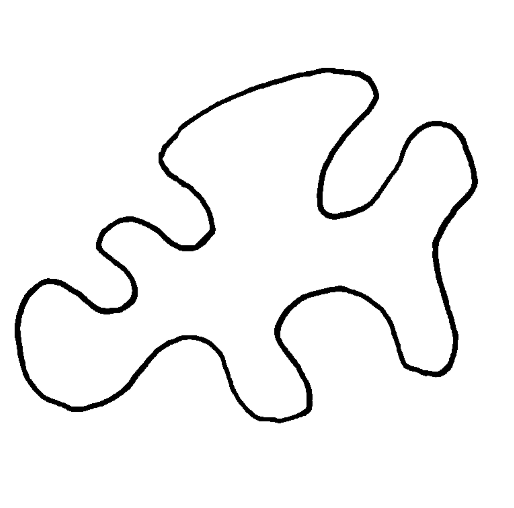} 
   \includegraphics[width=0.24\linewidth]{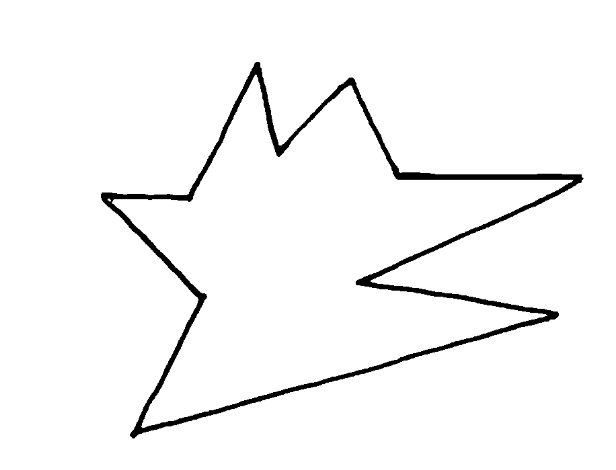} \hfill
   \includegraphics[width=0.24\linewidth]{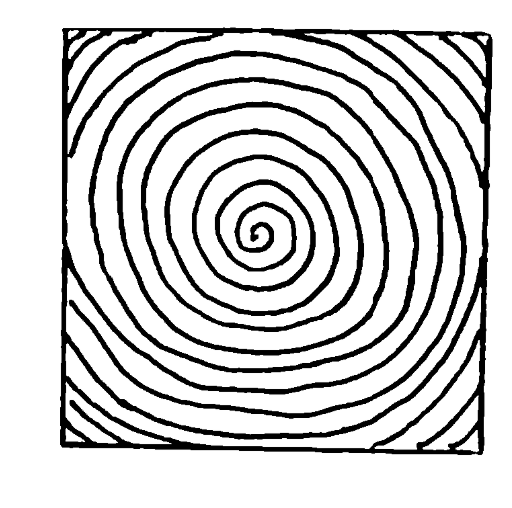} 
   \includegraphics[width=0.24\linewidth]{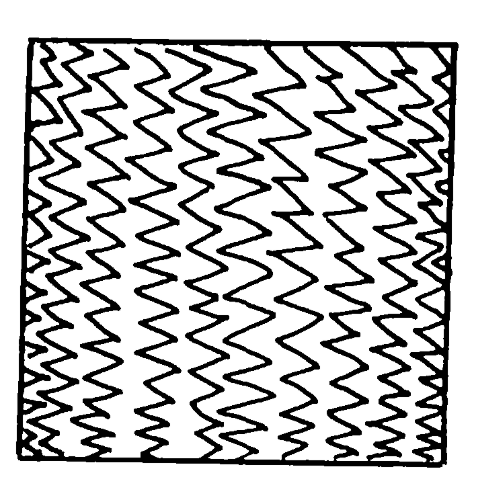}
   \includegraphics[width=0.24\linewidth]{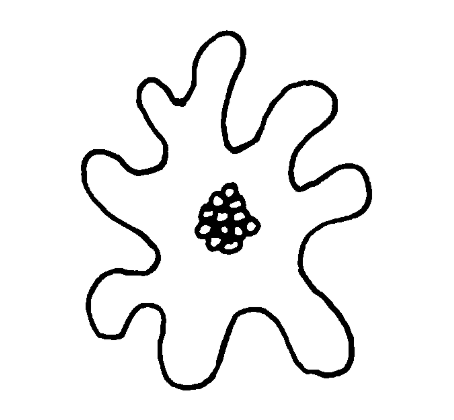}  
   \includegraphics[width=0.24\linewidth]{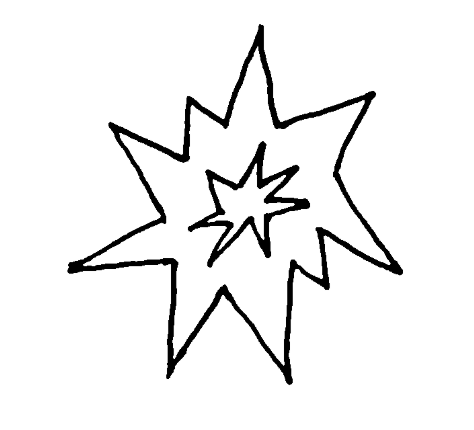} \hfill
   \caption{Image pairs from \citep{maurer2006shape}}
   \label{fig:maurer}
\end{figure}

\begin{figure}
    \centering
    \includegraphics[width=0.24\linewidth]{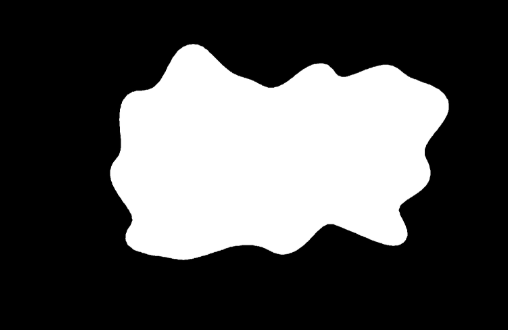}
    \includegraphics[width=0.24\linewidth]{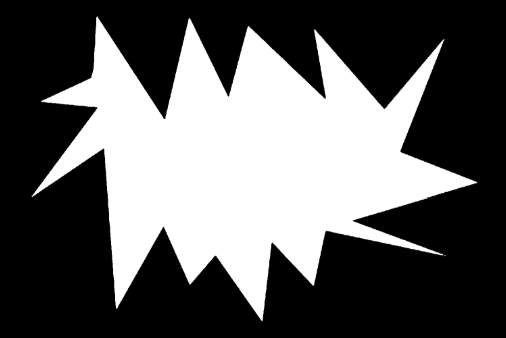}
    \includegraphics[width=0.24\linewidth]{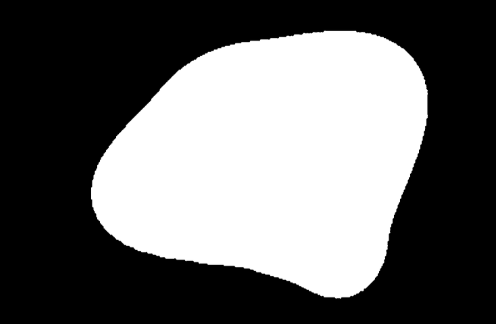} 
    \includegraphics[width=0.24\linewidth]{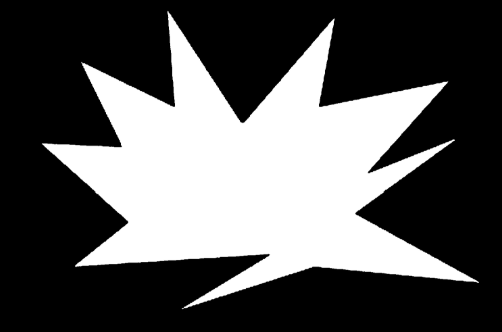} \hfill
    \includegraphics[width=0.24\linewidth]{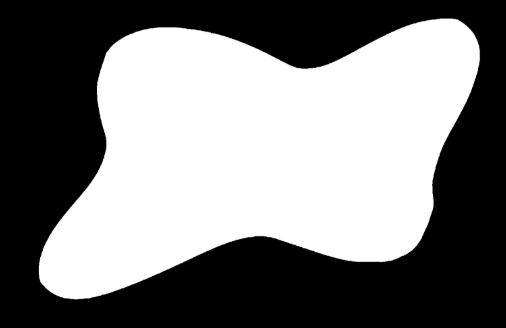}
    \includegraphics[width=0.24\linewidth]{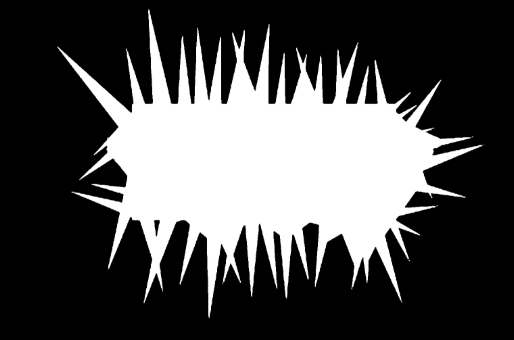}
    \includegraphics[width=0.24\linewidth]{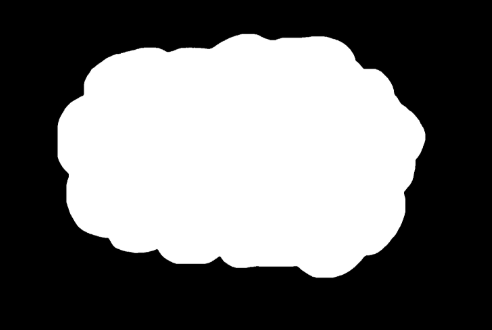}
    \includegraphics[width=0.24\linewidth]{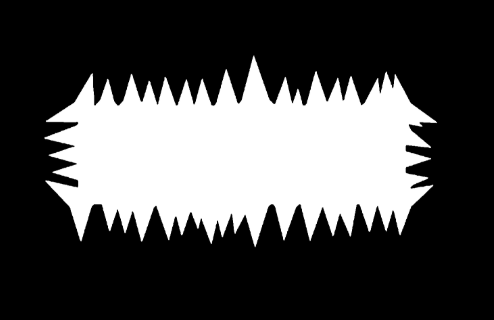} \hfill
    \caption{Image pairs from \citep{westbury2005implicit}}
    \label{fig:westbury}
\end{figure}

\begin{figure}
    \includegraphics[width=0.115\linewidth]{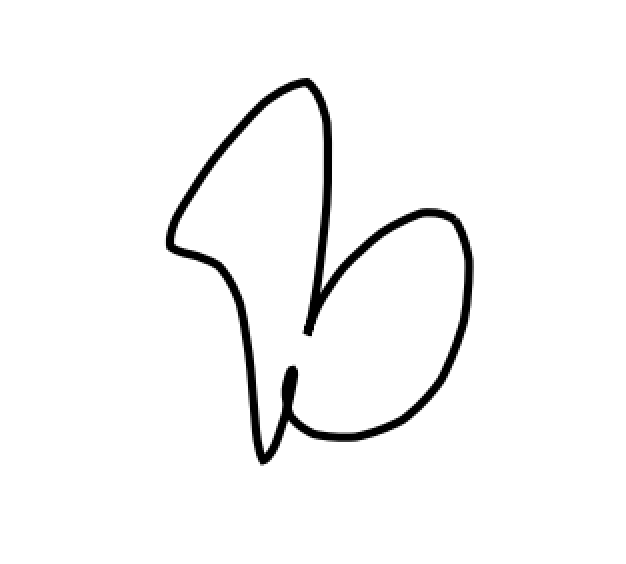}
    \includegraphics[width=0.115\linewidth]{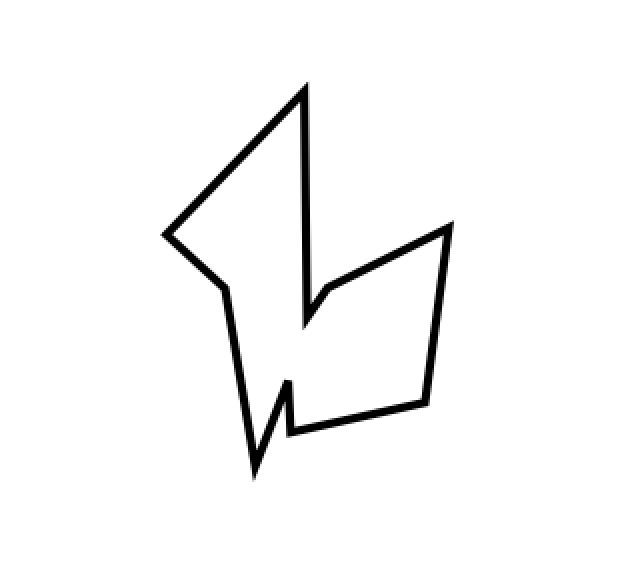}
    \includegraphics[width=0.115\linewidth]{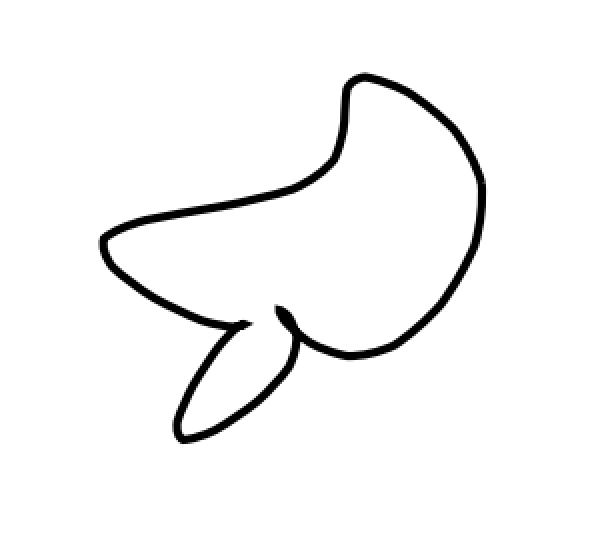}
    \includegraphics[width=0.115\linewidth]{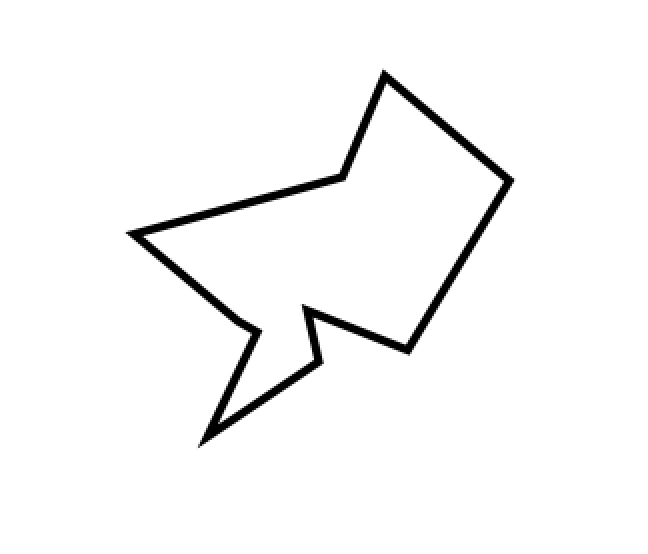}
    \includegraphics[width=0.115\linewidth]{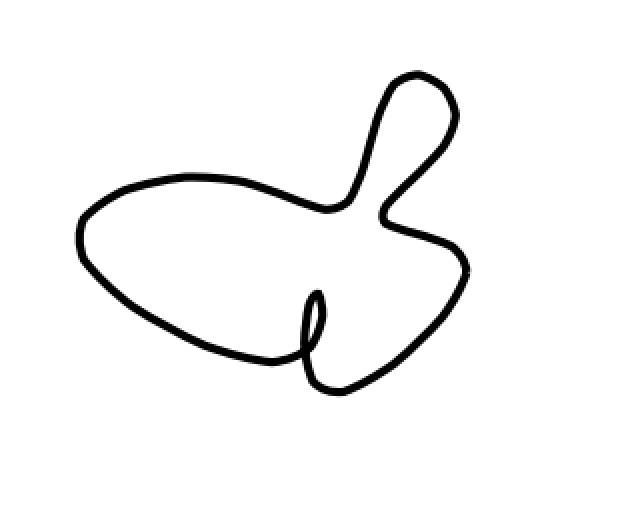}  
    \includegraphics[width=0.115\linewidth]{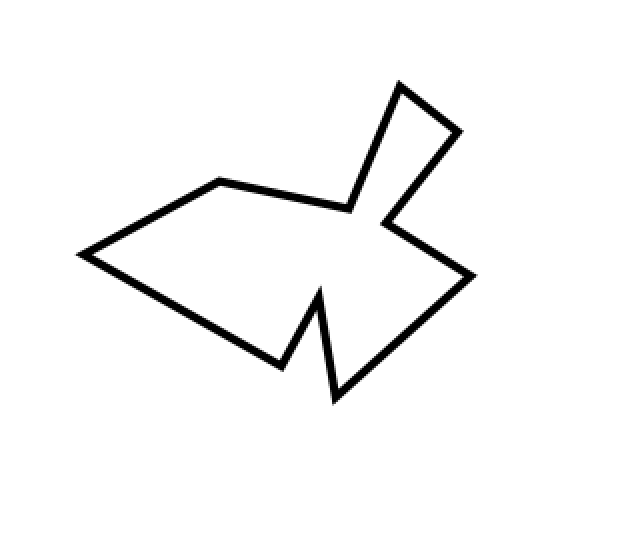}
    \includegraphics[width=0.115\linewidth]{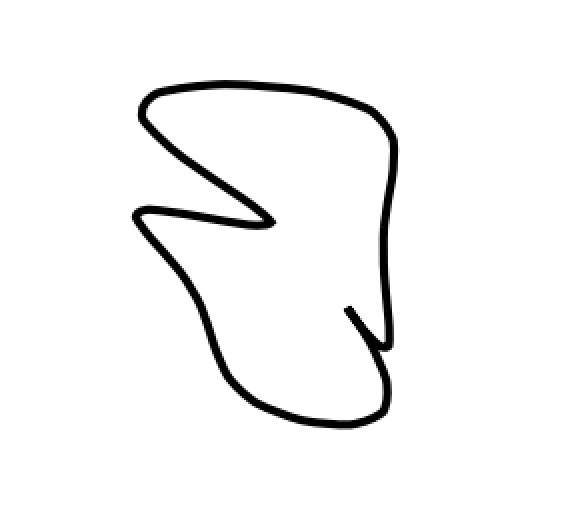}
    \includegraphics[width=0.115\linewidth]{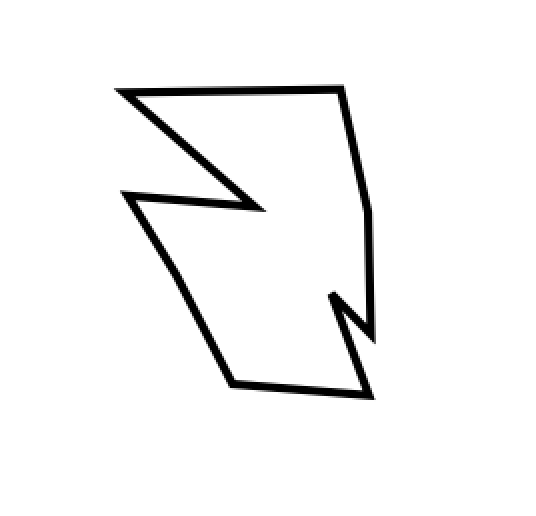} \hfill
    \includegraphics[width=0.115\linewidth]{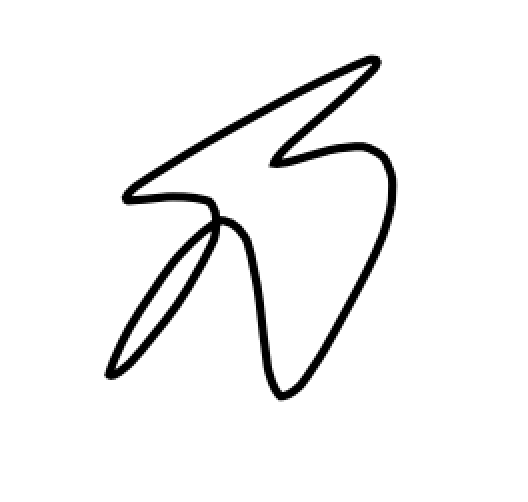} 
    \includegraphics[width=0.115\linewidth]{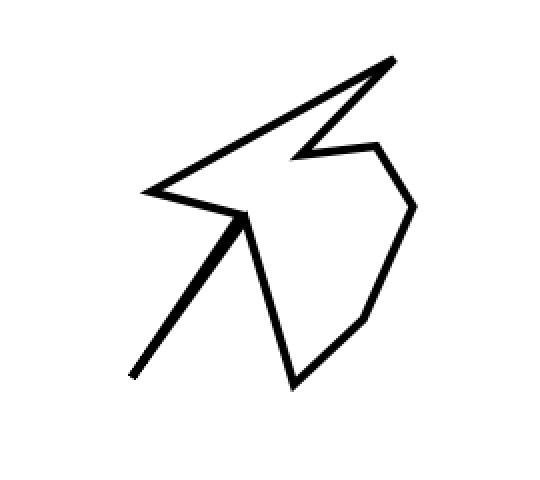}
    \includegraphics[width=0.115\linewidth]{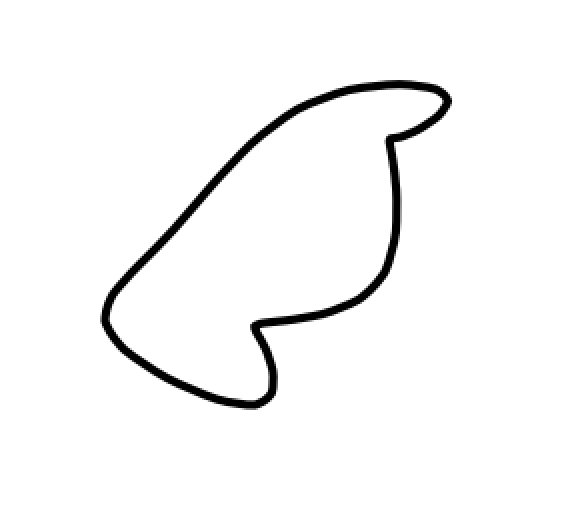} 
    \includegraphics[width=0.115\linewidth]{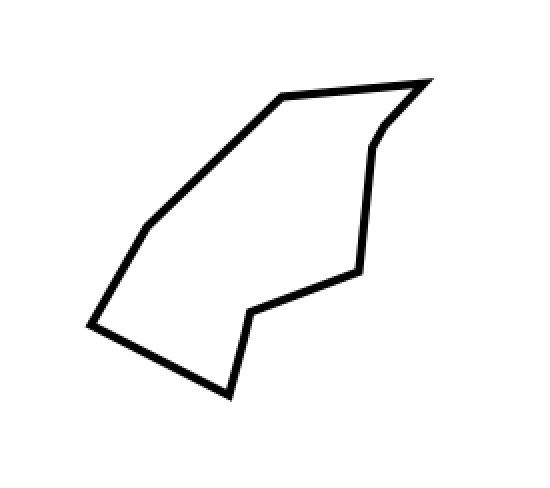}
    \includegraphics[width=0.115\linewidth]{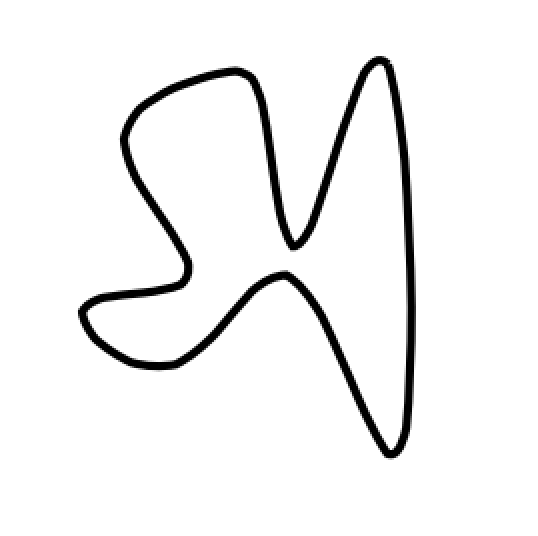} 
    \includegraphics[width=0.115\linewidth]{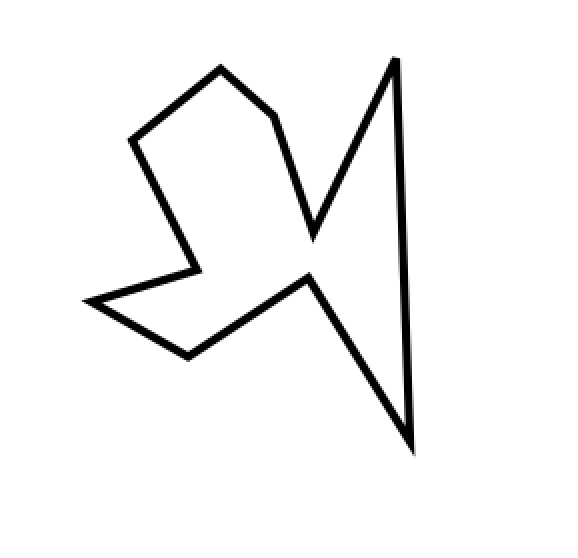}
    \includegraphics[width=0.115\linewidth]{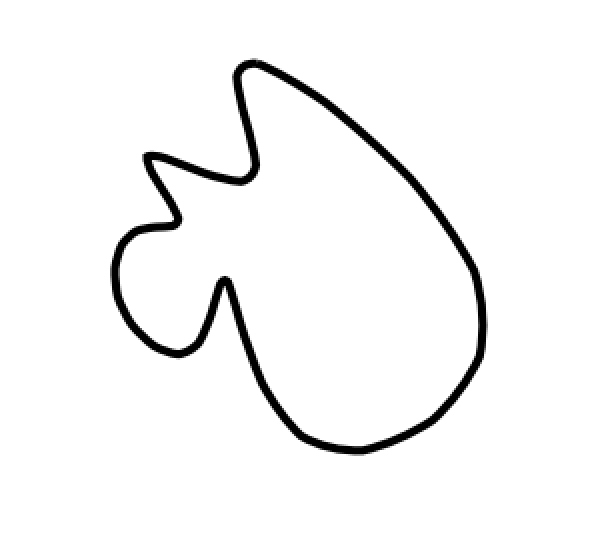}  
    \includegraphics[width=0.115\linewidth]{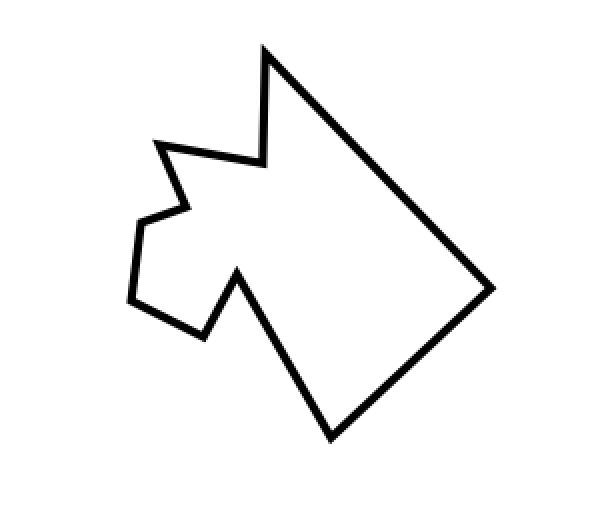} \hfill
  \caption{Newly generated image pairs}
  \label{fig:generated}
\end{figure}

The full set of images with visual shapes that were used in the experiments is shown here. Besides the original image pair from \citet{kohler1929gestalt, kohler1947gestalt}, we used four image pairs from \citet{maurer2006shape}, displayed in \autoref{fig:maurer}, four from \cite[][; \autoref{fig:westbury}]{westbury2005implicit}, and eight generated pairs following the method described by \citet{nielsen2013parsing} and used in \cite[][; \autoref{fig:generated};]{verhoef-etal-2024-kiki}. For each image pair, the Curved version is displayed on the left and the Jagged version on the right.

\subsection{Visual embeddings}\label{appendix:visual_embeddings}
\begin{figure}
    \centering
    \includegraphics[width=\linewidth]{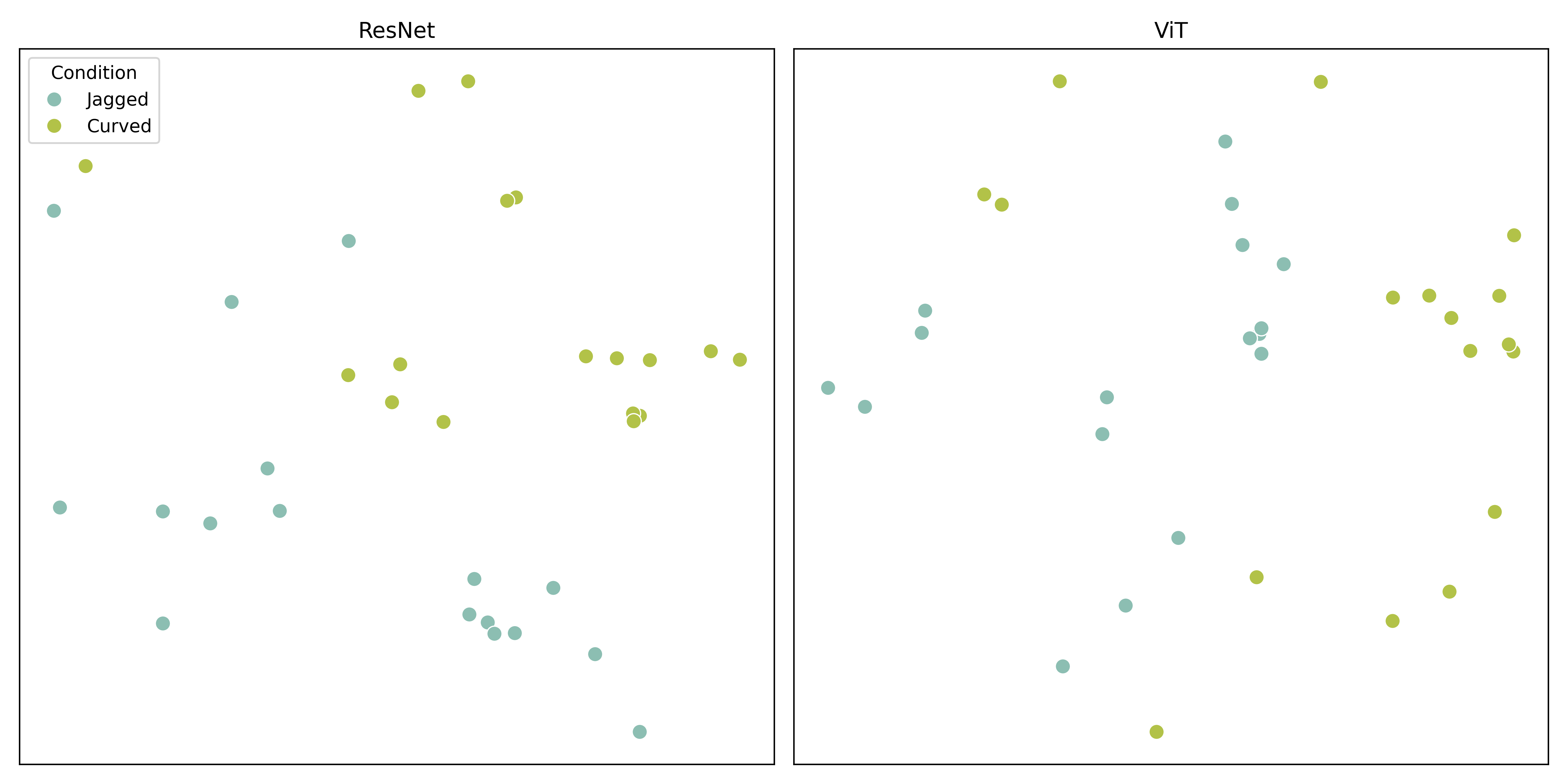}
    \caption{t-SNE plot showing how the vision models of different CLIP variants interpret images from conditions. The colour shades indicate which image condition an image belongs to. In order to correctly match labels to an image, a model must be able to discriminate the images based on their shape. This is clearly possible.}
    \label{appendix:fig:tsne-image}
\end{figure}

Similar to the textual embeddings presented before, \autoref{appendix:fig:tsne-image} displays a t-SNE visualisation of the models' visual embeddings. It is visible that the models should, in principle, be able to disambiguate our target images based on their condition, even when they only differ in how the randomly distributed points are connected.

\section{Unique labels}\label{appendix:ratio_unique_labels}
To test whether the models in the first experiment (\autoref{exp1}) change their predictions according to a set of images, we report the ratio of unique labels in \autoref{appendix:tab:ratio_unique}.
In this case, we are not concerned with whether the models make predictions that align with the bouba-kiki effect, but rather with the diversity of their predictions.
We calculate the ratio of unique labels for each word type across all images and the ten prompts.
These ratios should, in a scenario where cross-modal associations are present, be high, and the number of correct trials, as seen in \autoref{fig:probability_winners}, would be above chance. 
However, it is clear that the models somewhat change their predictions (mainly for the initial pseudowords) when they are conditioned on different images, but mostly \textit{collapse} onto the same labels for different images. 
\autoref{appendix:tab:ratio_unique} confirms this by showing that there is only slight variation among picked labels for each prompt. 
Given that responses are only counted as congruent when both predictions (i.e. the prediction for a jagged and curved shape) are correct, this explains why we find that both models perform at chance levels. 
A qualitative example is provided in \autoref{appendix:tab:example_unique}, which shows the unique labels and their ratios for a single prompt.
The model and word type that are most affected by different images (i.e. have a high ratio of unique words) also exhibit the strongest bouba-kiki like effect \autoref{fig:probability_winners}.
This appears to happen even though, in the case of both ViT and ResNet, and \citeauthor{alper2024kiki} labels, the majority of the labels would be associated with sharp images by humans.  
% alper restnet 12 out of 16
% alper vit 17 23

\begin{table}
    \centering
    \begin{tabular}{llc}
    \toprule
        Model   & Word type             & Ratio \\\midrule
        ResNet  & Initial words         & .675  \\
        ResNet  & English adj.          & .275  \\
        ResNet  & Nielsen et al.        & .329  \\
        ResNet  & Alper et al.          & .444  \\\midrule
        ViT     & Initial words         & .800  \\
        ViT     & English adj.          & .320  \\
        ViT     & Nielsen et al.        & .331  \\
        ViT     & Alper et al.          & .450  \\
    \bottomrule
    \end{tabular}
    \caption{The average ratio of unique labels chosen for each image (n=34) across different prompts (n=10) in different label sets. The latter means that we divide the unique labels by the length of the set of possible labels to gauge diversity. A high ratio indicates that the corresponding model assigned different labels to different images.}
    \label{appendix:tab:ratio_unique}
\end{table}

\begin{table}[hb]
    \centering
    \begin{tabular}{llll}
    \toprule
        Model   & Word type             & Ratio & Uniquely chosen labels                                                                                                                \\\midrule
        ResNet  & Initial words         & .250  &\makecell[tl]{bouba}                                                                                                                   \\
        ResNet  & English adj.          & .300  &\makecell[tl]{angular, circular, curved, prickly, rotund, spiky}                                                                       \\
        ResNet  & Nielsen et al.        & .412  &\makecell[tl]{kaykee, kuhpay, kuhpee, kuhpuh, lahmoo, lohloh, lohmah,\\ lohmoo, mohmah, nahmoo, nohmoo, paykuh, peepay, teepee}        \\
        ResNet  & Alper et al.          & .471  &\makecell[tl]{kehake, kehike, ladula, lunulu, malama, nulunu, paxapa,\\ pihapi, pikepi, pikipi, pixapi, sepise, tatata, tekete, xaxixa,\\ xehexe}                                                                                                                                                                                    \\\midrule
        ViT     & Initial words         & .500  &\makecell[tl]{bouba, takete}                                                                                                           \\
        ViT     & English adj.          & .250  & \makecell[tl]{angular, circular, curved, rotund, spiky}                                                                               \\
        ViT     & Nielsen et al.        & .412  & \makecell[tl]{keekay, kuhtay, lahmoo, lohlah, loomoh, mahloh, mahnoh,\\ mohloo, mohnoh, nohloh, noonoo, taypay, taypuh, teepee}       \\
        ViT     & Alper et al.          & .676  & \makecell[tl]{dododo, hexehe, kixiki, lamola, lubalu, lunulu, mamuma,\\ monomo, mubomu, patipa, pisipi, pixapi, sahisa, satasa,\\ tehite, texate, xahexa, xakixa, xasixa, xehexe, xikexi, xipaxi,\\ xisixi}                                                                                                                         \\
    \bottomrule
    \end{tabular}
    \caption{The set of unique labels chosen across all images (n=34) and its ratio for an example prompt (`The label for this image is <label>'). The latter means that we divide the unique labels by the length of the set of possible labels to gauge diversity. A high ratio indicates that the corresponding model assigned different labels to different images.}
    \label{appendix:tab:example_unique}
\end{table}

\newpage

\section{Grad-Cam visualisations, consistency, and additional results}\label{appendix:gradcam}
\begin{figure}[h]
  \centering
  \includegraphics[width=\linewidth]{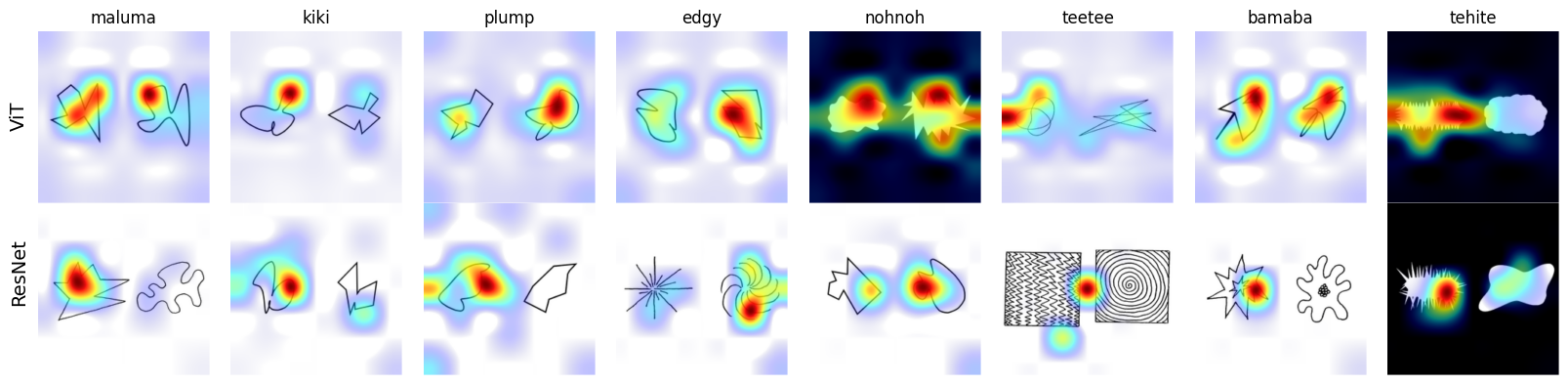}
  \caption{Visualising the attention pattern for the text prompt: `The label for this image is <label>' in both models.}
  \label{fig:attention_pattern_example}
\end{figure}

Figure \ref{fig:attention_pattern_example} provides example visualisations of the attention patterns for different randomly selected image pairs and randomly selected labels across both models. 
Here, we sampled different image pairs for the models as to give a complete view of the images used.
The pseudowords are consistent for the columns.
The images reveal that models mostly attend to the centres of shapes and/or focus on non-informative background areas.
The latter is also described by \citet{darcet2024vision}.
Although this is a small sample, it is clear that the target labels do not steer models towards shape-specific features.

\subsection{Prediction consistency}
To test whether the models in the second experiment (\autoref{exp2}) change their predictions resulting from different shape positions, we report the ratio of consistency in \autoref{appendix:tab:ratio_consistent}.
In this case, we are not interested in whether the models make predictions that align with the bouba-kiki effect, but rather in the consistency of their predictions.
Both models are rather consistent in the mappings they make between labels and shapes.

\begin{table}[b]
    \centering
    \begin{tabular}{cccc}
    \toprule
        Model   & Word type             & Ratio \\ \midrule
        ResNet  & Initial words         & .799  \\
        ResNet  & English adjectives    & .761  \\
        ResNet  & Nielsen et al.        & .758  \\
        ResNet  & Alper et al.          & .765  \\ \midrule
        ViT     & Initial words         & .723  \\
        ViT     & English adjectives    & .749  \\
        ViT     & Nielsen et al.        & .722  \\
        ViT     & Alper et al.          & .739  \\
    \bottomrule
    \end{tabular}
    \caption{The ratio of consistently attending to the same shape in a different position when presented with the same label. Values are averages across image pairs, prompts, and labels. A high ratio indicates that the corresponding model consistently focuses on the same shape, even when the target location is different.}
    \label{appendix:tab:ratio_consistent}
\end{table}

\subsection{Quantifying model preference}
The results presented in \autoref{exp2} utilise the sum of attention values to quantify the models' preferences, as this aligns with humans' holistic perception \citep{Taubert2011TheRO, Zhao2016hollisticprocessing, wong2011wordshollistic}.
Though artificial models may show their preference differently.  
To this end, we additionally experimented with the entropy and centroid-of-attention as quantifiers of model preference. 

Comparing the predictions (averaged over all images, labels, and prompts) quantified by the centroid of attention with the sum of attention, we find that the predictions using the centroid of attention strongly overlap with those resulting from the sum of attention (ViT: $85.6\%$ and ResNet: $88.0\%$). 
Given this considerable overlap, it is not surprising that the results remain highly similar when using this alternative method (\autoref{appendix:tab:correctness}.
Yet, for the entropy of attention, the predictions (where the image half with the lower entropy acts as the model’s preference) for ViT overlap strongly ($80.2\%$) but not for ResNet ($21.1\%$). 
% Overlap   Sum-Centr   sum_entropy
% ResNet    0.880195	0.211017
% ViT	    0.856298	0.802405
This entropy method, despite yielding different predictions, still results in a slightly lower number of correctly predicted images, with only $48.8\%$ of the shapes correctly identified, compared with $51.9\%$ for the quantification method that uses the sum of attention. 
So if we were to use entropy-based predictions, the performance would be even worse than predictions based on the sum of attention.

\begin{table}
    \centering
    \begin{tabular}{cccc}
    \toprule
        Model   & Sum   & Entropy   & Centroid  \\ \midrule
        ResNet  & .519  & .488      & .515      \\
        ViT     & .522  & .514      & .515      \\
    \bottomrule
    \end{tabular}
    \caption{The correctness of all model prediction types using different quantifications averaged across all images, labels, and prompts.}
    \label{appendix:tab:correctness}
\end{table}

%         Correct	    Correct_entropy	   Correct_centroid
% Model			
% ResNet	0.518620	0.488433	       0.515404
% ViT	    0.522175	0.514085           0.515446

\end{document}